\pdfoutput=1

\documentclass[11pt]{article}

\IfFileExists{acl.sty}{\usepackage[preprint]{acl}}{\usepackage[preprint]{latex/acl}}

\usepackage{times}
\usepackage{latexsym}

\usepackage[T1]{fontenc}
\usepackage[utf8]{inputenc}
\usepackage{microtype}
\usepackage{inconsolata}

\usepackage{graphicx}
\graphicspath{{imgs/}{imgs/}}
\usepackage{subcaption}
\usepackage{pdfpages}

\usepackage{booktabs}
\usepackage{multirow}
\usepackage{array}
\usepackage{colortbl}
\usepackage{makecell}
\usepackage{siunitx}
\usepackage{adjustbox}

\usepackage{amsmath,amssymb}

\usepackage{xcolor}
\definecolor{rowgray}{gray}{0.93}
\definecolor{teal}{rgb}{0.0,0.50,0.45}
\definecolor{lightgray}{gray}{0.9}
\newcommand{\grayrow}{\rowcolor{lightgray}}

\usepackage{algorithm}
\usepackage{algpseudocode}

\usepackage{tikz}
\usetikzlibrary{arrows.meta, positioning, shapes.multipart, shapes.geometric, calc, patterns, decorations.markings}

\usepackage{listings}
\usepackage{acro}

\newcommand{\gemma}{Gemma}
\newcommand{\llama}{LLaMA}
\newcommand{\qwen}{Qwen}

\DeclareAcronym{llm}{short=LLM,long=Large Language Model}
\DeclareAcronym{sr}{short=SR,long=Symbolic Regression}
\DeclareAcronym{icp}{short=ICP,long=Iterative Closest Point}
\DeclareAcronym{nlp}{short=NLP,long=Natural Language Processing}
\DeclareAcronym{lb}{short=LB,long=Lemniscates of Bernoulli ($\infty$)}
\DeclareAcronym{da}{short=$\mathbb{D}a$,long=Designer Agent}
\DeclareAcronym{ca}{short=$\mathbb{C}a$,long=Critique Agent}
\DeclareAcronym{llama}{short=Llama,long=Llama3.3:70B}
\DeclareAcronym{gemma}{short=Gemma,long=Gemma3:12B}
\DeclareAcronym{qwen}{short=Qwen,long=Qwen3:4B}
\DeclareAcronym{lrm}{short=LRM,long=Large Reasoning Model}

\title{Symbolic Intermediaries as a Linguistic--Numerical Interface\\for LLM-Driven Geometric Reasoning}

\author{
Jo\~{a}o Pedro Gandarela\textsuperscript{1,2} \quad
Thiago Rios\textsuperscript{3} \quad
Stefan Menzel\textsuperscript{3} \quad
Andr\'{e} Freitas\textsuperscript{1,4,5} \\
\textsuperscript{1}Idiap Research Institute, Switzerland \quad
\textsuperscript{2}\'{E}cole Polytechnique F\'{e}d\'{e}rale de Lausanne (EPFL), Switzerland \\
\textsuperscript{3}Honda Research Institute Europe, Germany \\
\textsuperscript{4}Department of Computer Science, University of Manchester, UK \\
\textsuperscript{5}National Biomarker Centre, CRUK-MI, University of Manchester, UK \\
\texttt{firstname.lastname@idiap.ch} \quad
\texttt{firstname.lastname@honda-ri.de}
}

\begin{document}
\maketitle

\begin{abstract}
\noindent
Large Language Models (LLMs) display reasoning capabilities over linguistic and symbolic objects but have limited capabilities to directly interpret the continuous numerical outputs of physics simulators, e.g., distances, curvatures, and trajectories that resist discrete tokenisation. Across spatially grounded engineering reasoning tasks, from mechanism design to motion planning, this defines a fundamental gap, which limits the wider application of LLMs within broader geometrical domains, for exmaple interfacing with physics simulators.
We propose \emph{symbolic intermediaries}, compact analytical expressions discovered via symbolic regression, as a structured interface that translates a simulator's numerical traces into a symbolic form, which language models can interpret, compare, and critique while preserving the original geometric semantics.
Around this interface we build an agentic coordination-and-refinement loop: a \emph{design agent} maps natural-language specifications to executable simulation code, a \emph{critique agent} reasons over the shared symbolic vocabulary, and a revision step turns this feedback into grounded refinement decisions, enabling inference-time generalization without parameter updates.
On the MSynth benchmark for planar mechanism synthesis, all three evaluated LLM agents outperform a budget-matched genetic-algorithm baseline by 19--53\% (up to 63\% lower median error with feedback), and analysis of the critique entries across three model architectures shows that the interface shifts reasoning from generic structural commentary to grounded geometric verification.
The principle of translating continuous simulation outputs into symbolic forms generalises to any domain where simulator behaviour must be interpreted linguistically.
\end{abstract}

\section{Introduction}
\label{sec:intro}

\ac{llm}s show strong capabilities in symbolic and linguistic reasoning, from generating code to proving mathematical theorems \cite{fu2024limsimclosedloopplatformdeploying, bhat2024groundingllmsrobottask,pmlr-v205-ichter23a,wei2022cot,kojima2022zeroshot,trinh2024alphageometry}. Yet a broad class of scientific and engineering problems, from molecular geometry and protein structure prediction to robot motion planning and structural design, is governed by \emph{continuous spatial relationships}, such as distances, angles, curvatures, and trajectories that evolve smoothly through physical space, rather than by symbol manipulation.

Such problems are resolved by \emph{physics simulators}, which numerically integrate the governing equations and evaluate candidate designs with high fidelity, but their outputs (coordinate traces, distance fields, stress tensors) are continuous, high-dimensional, and resist the discrete tokenisation on which language models rely, creating a representational mismatch that limits the direct application of \ac{llm}s to spatially grounded tasks \cite{wang2024gensim,rana2023sayplan,mouselinos-etal-2024-beyond}. The fundamental bottleneck is therefore the absence of a structured \textbf{interface} between the linguistic domain of the \ac{llm} and the spatial, simulator-governed domain in which these problems are actually solved, rather than a deficiency of either system in isolation.

To bridge these two levels of reasoning, we propose a method built around a \textbf{symbolic intermediary}: a compact analytical representation, discovered through \ac{sr}, that translates the simulator's continuous spatial output into a symbolic form that \ac{llm}s can interpret, compare, and critique, while preserving the geometric semantics of the original simulation. An expression such as $y = a\sin(bx + c) + d$ makes periodicity, amplitude, and phase explicit in a vocabulary that is simultaneously symbolic (amenable to linguistic reasoning) and geometric (faithful to the simulator's output).

This makes the approach inherently \emph{neuro-symbolic}: linguistic generation by the \ac{llm} is coupled to a symbolic representation grounded in numerical simulation, yielding the rigour and verifiability that purely linguistic reasoning lacks and that has proven decisive in formal-reasoning systems where correctness must be guaranteed rather than approximated \cite{hubert2025olympiad}. Situated between the two, the intermediary closes an \emph{agentic} reasoning loop: the \ac{llm} proposes design hypotheses, the simulator evaluates them numerically, the intermediary translates those numerical results into symbolic form, and the \ac{llm} reasons over the symbolic feedback to propose refinements \cite{shinn2023reflexion,park2023generative,wang2024agentsurvey}.

We demonstrate this interface paradigm on \emph{planar mechanism synthesis}, a canonical engineering task in which an assembly of rigid links and revolute joints must produce a prescribed end-effector trajectory (Fig.~\ref{fig:problem}). Concretely, a four-bar linkage has two ground-fixed pivots and a moving \emph{coupler point}; as the crank rotates, that point traces a closed 2D curve, and the design problem is to choose link lengths so this curve matches a target shape.

Mechanism synthesis is an ideal test bed for two reasons. First, the simulator supplies \emph{ground-truth rigour}: the coupler trajectory is fully determined by the kinematic equations, so a design is right or wrong by construction rather than by subjective judgement, which is exactly the property that lets a symbolic intermediary deliver verifiable feedback. Second, the task still demands genuine \emph{multi-representational reasoning}: a solution must coordinate a linguistic specification, a discrete topology, continuous kinematic constraints, and a 2D motion path, while existing automated methods require expert-driven setup and lack scalability \cite{Garc_a_Marina_2020, Tsai1999, Romero2019, vasiliu2001dimensional, sonntag2024machine}. The interface principle is not specific to mechanisms: it could extend to other domains where a simulator produces continuous numerical behaviour that must be interpreted linguistically, which we leave to future work.

\definecolor{cbBlue}{HTML}{0072B2}
\definecolor{cbOrange}{HTML}{E69F00}
\definecolor{cbVermillion}{HTML}{D55E00}
\definecolor{cbGrayL}{HTML}{BFBFBF}

\begin{figure*}[t]
  \centering
  \begin{tikzpicture}[
      >=Stealth, font=\small,
      joint/.style   ={circle, draw, fill=white, line width=0.7pt, inner sep=1.6pt},
      link/.style    ={line width=1.1pt, line cap=round},
      crank/.style   ={line width=2.0pt, cbBlue, line cap=round},
      ghost/.style   ={line width=0.7pt, cbGrayL, dashed, line cap=round},
      ground/.style  ={pattern=north east lines},
      panel/.style   ={draw=black!55, rounded corners, fill=black!4,
                        inner sep=6pt, align=left},
      pathline/.style={line width=1.4pt, cbOrange,
                        dash pattern=on 4pt off 2pt on 1pt off 2pt,
                        postaction={decorate},
                        decoration={markings,
                          mark=at position 0.58 with {\arrow{Stealth[length=4pt]}}}},
      flow/.style    ={->, line width=1pt, black!65}]

    \node[panel] (setrep) at (0,0) {%
      $\mathcal{M}=(\mathcal{J},\mathcal{L})$\\[2pt]
      $\mathcal{J}=\{A,B,C,D,E\}$\\[1pt]
      $\mathcal{L}=\{\,\overline{AB},\overline{BC},\overline{CD},\overline{BE},\overline{CE}\,\}$\\[1pt]
      fixed: $A,D$ \;\; target: $\mathcal{T}$};
    \node[anchor=south, font=\small\bfseries] at (setrep.north) {(b) set/topology};

    \begin{scope}[shift={(6.0,-0.55)}, scale=0.9]
      \coordinate (A) at (0,0);          
      \coordinate (D) at (3.0,0);        
      \coordinate (Em)  at (2.55,2.15);  
      \coordinate (Eg1) at (0.85,2.95);  
      \coordinate (Eg2) at (1.45,1.35);  
      \draw[pathline]
        plot[smooth cycle, tension=0.75] coordinates
        {(0.55,2.05) (Eg1) (1.55,3.25) (2.25,2.95) (Em)
         (2.15,1.55) (Eg2) (0.75,1.55)};
      \node[cbOrange, font=\footnotesize] (pathlbl) at (1.55,3.55)
        {coupler path $=\mathcal{T}$};

      \draw[dotted, cbBlue!70] (A) circle (0.80);
      \draw[->, cbBlue, line width=1pt]
        ($(A)+(-25:1.02)$) arc[start angle=-25, end angle=210, radius=1.02];
      \node[cbBlue, font=\footnotesize] at (-0.55,-0.85) {crank rotates $360^\circ$};

      \foreach \g in {A,D}{%
        \draw[ground] ($(\g)+(-0.42,-0.42)$) rectangle ($(\g)+(0.42,-0.10)$);
        \draw[line width=0.7pt] ($(\g)+(-0.42,-0.10)$)--($(\g)+(0.42,-0.10)$);}

      \coordinate (Bg1) at ($(A)+(110:0.80)$); \coordinate (Cg1) at ($(D)+(150:1.70)$);
      \draw[ghost] (A)--(Bg1)--(Cg1)--(D); \draw[ghost] (Bg1)--(Eg1)--(Cg1);
      \coordinate (Bg2) at ($(A)+(18:0.80)$);  \coordinate (Cg2) at ($(D)+(102:1.70)$);
      \draw[ghost] (A)--(Bg2)--(Cg2)--(D); \draw[ghost] (Bg2)--(Eg2)--(Cg2);

      \coordinate (B) at ($(A)+(52:0.80)$);   
      \coordinate (C) at ($(D)+(122:1.70)$);  
      \draw[crank] (A)--(B);                              
      \draw[link]  (B)--(C);                              
      \draw[link]  (C)--(D);                              
      \draw[link]  (B)--(Em) -- (C);                      
      \foreach \p in {A,B,C,D}{\node[joint] at (\p){};}
      \node[diamond, draw, fill=cbVermillion, inner sep=2pt] at (Em){};

      \node[below=3pt] at (A) {$A$}; \node[below=3pt] at (D) {$D$};
      \node[left=3pt]  at (B) {$B$}; \node[right=2pt] at (C) {$C$};
      \node[right=3pt, cbVermillion] at (Em) {$E$ (coupler point)};
      \node[cbBlue, font=\footnotesize, anchor=west]
        at ($(A)!0.5!(B)+(0.12,0.10)$) {crank};
    \end{scope}
    \node[anchor=south, font=\small\bfseries]
      at ($(pathlbl.north)+(0,0.05)$) {(a) geometric/physical};

    \node[panel] (symrep) at (12.7,0) {%
      $y = a\sin(bx + c) + d$\\[3pt]
      \footnotesize periodicity, amplitude,\\[-1pt]
      \footnotesize phase made explicit};
    \node[anchor=south, font=\small\bfseries] at (symrep.north) {(c) symbolic};

    \draw[flow] (setrep.east) -- node[above, font=\footnotesize]{simulate}
                ($(setrep.east)+(1.9,0)$);
    \draw[flow] ($(symrep.west)-(1.9,0)$) -- node[above, font=\footnotesize]{\ac{sr}}
                (symrep.west);
  \end{tikzpicture}
  \caption{One planar four-bar mechanism in three coordinated representations: the
  designer agent emits the discrete \emph{set/topology} form (b); the simulator turns
  it into the \emph{geometric} view (a), whose coupler path is the prescribed target
  $\mathcal{T}$; symbolic regression compresses that path into the \emph{symbolic}
  form (c), the shared vocabulary the critique agent reasons over.
  }
  \label{fig:problem}
\end{figure*}

\paragraph{Contributions.}
To our knowledge, this is the first method to couple an agentic \ac{llm} design--critique--revision loop with a physics simulator through a symbolic interface traversed in \emph{both directions}. Concretely:
\textbf{(1)} \emph{An agentic coordination-and-refinement loop.} A design agent and a critique agent coordinate around a shared symbolic representation, and a revision step closes the loop: at each iteration it selects which geometric deficiency to correct, grounds that choice in the symbolic feedback, and commits valid designs to an external memory. This realises geometric design as learning (in Mitchell's sense~\cite{mitchell1997machine}) without gradients, labels, or fine-tuning (\S\ref{rq:1},~\ref{rq:2}).
\textbf{(2)} \emph{Symbolic intermediaries as the enabling interface.} We introduce symbolic intermediaries, compact analytical expressions, as a bidirectional bridge between the linguistic domain of \ac{llm}s and the spatial domain of physics simulators, lifting simulator output to symbol and translating symbolic critique back to executable design. Systematic analysis of the critique entries shows the interface changes \emph{what} critiques reason over, increasing engagement with discovered symbolic structure across all models and, in reasoning-trained models, shifting reasoning toward grounded geometric verification, a distinction of grounding rather than raw frequency that aggregate metrics obscure (\S\ref{rq:3}).
\textbf{(3)} \emph{Benchmark and evidence.} We introduce \textit{MSynth}\footnote{anonymized}, a benchmark for planar mechanism synthesis, and show that the method achieves up to 90\% improvement in geometric fidelity and statistically significant gains over genetic-algorithm baselines, with systematic ablations revealing how symbolic representations interact with model capacity (\S\ref{rq:4}).

\section{Related Work}
\label{sec:related_work}

\paragraph{Spatial reasoning, tool use, and self-refinement.}
Language models struggle with mental rotation, 3D scene understanding, and continuous geometric inference \cite{wang2024gensim,rana2023sayplan}, which prior work addresses through visual grounding \cite{pmlr-v205-ichter23a}, code generation for simulation \cite{fu2024limsimclosedloopplatformdeploying}, or specialised spatial training. Rather than augmenting spatial perception, we translate the geometric problem into a symbolic representation that aligns with the model's existing reasoning strengths. This builds on in-context learning \cite{brown2020language,garg2022incontext} and tool-augmented LMs \cite{schick2023toolformer,qin2024toolllm,gao2023pal}, with tools here acting as \emph{representational translators} rather than mere information sources. Iterative self-refinement \cite{dziri2023faith,madaan2023selfrefine,huang-chang-2023-towards,shinn2023reflexion,wang2023selfconsistency} and multi-agent plan/critique loops \cite{xiang-etal-2024-retrospex,shang2025agentsquare,yang2025agentoccam,hu2024hiagenthierarchicalworkingmemory,yao2023react,wu2024agentkit,gurtler2021hierarchical,yao2023tot,park2023generative,wang2024agentsurvey,li2023camel} provide closely related machinery; our neuro-symbolic component is distinctive because \ac{sr} expressions are \emph{discovered from data} rather than hard-coded, producing a dynamic symbolic layer that adapts per problem.

\paragraph{Mechanism Synthesis and Data-Driven Design}
One-shot neural generators can map target trajectories to linkage parameters \cite{articlebai_Shaoping_Angeles_Jorge,lee2024deep}, and gradient-based methods can further optimise topology and dimensions, though they remain sensitive to initialisation and penalty weighting \cite{singh2024new}. Large curated datasets (e.g., LINKS, ASME) provide static corpora for retrieval or one-shot learning \cite{nobari2022links,nobari2024linklearningjointrepresentations,nurizada2021dataset,venkataraman2018learningkinematicdescriptionsusing,martínmartín2018rbodatasetarticulatedobjects}. In contrast, our approach iteratively generates candidates and synthesises symbolic and numeric critiques to steer the search, without relying on pre-computed databases or brittle penalty tuning.
\section{Method: A Linguistic--Numerical Interface for Geometric Reasoning}
\label{sec:method}

We instantiate symbolic intermediaries as an interface layer on planar mechanism synthesis: given analytical equations $\mathcal{T}$ defining a desired end-effector trajectory in $\mathbb{R}^{2}$, the method must produce a mechanism whose motion path $\mathcal{G}$ minimises $d(\mathcal{T},\mathcal{G})$ subject to structural constraints. Conventional approaches use kinematic graphs and numerical optimisation but require expert-driven setup and lack scalability; we propose a neuro-symbolic \cite{NAWAZ2025200541,wan2024cognitiveaisystemssurvey} alternative in which the \ac{llm} and simulator communicate through a shared symbolic interface.

\begin{figure*}[t]
    \centering
    \includegraphics[width=2\columnwidth]{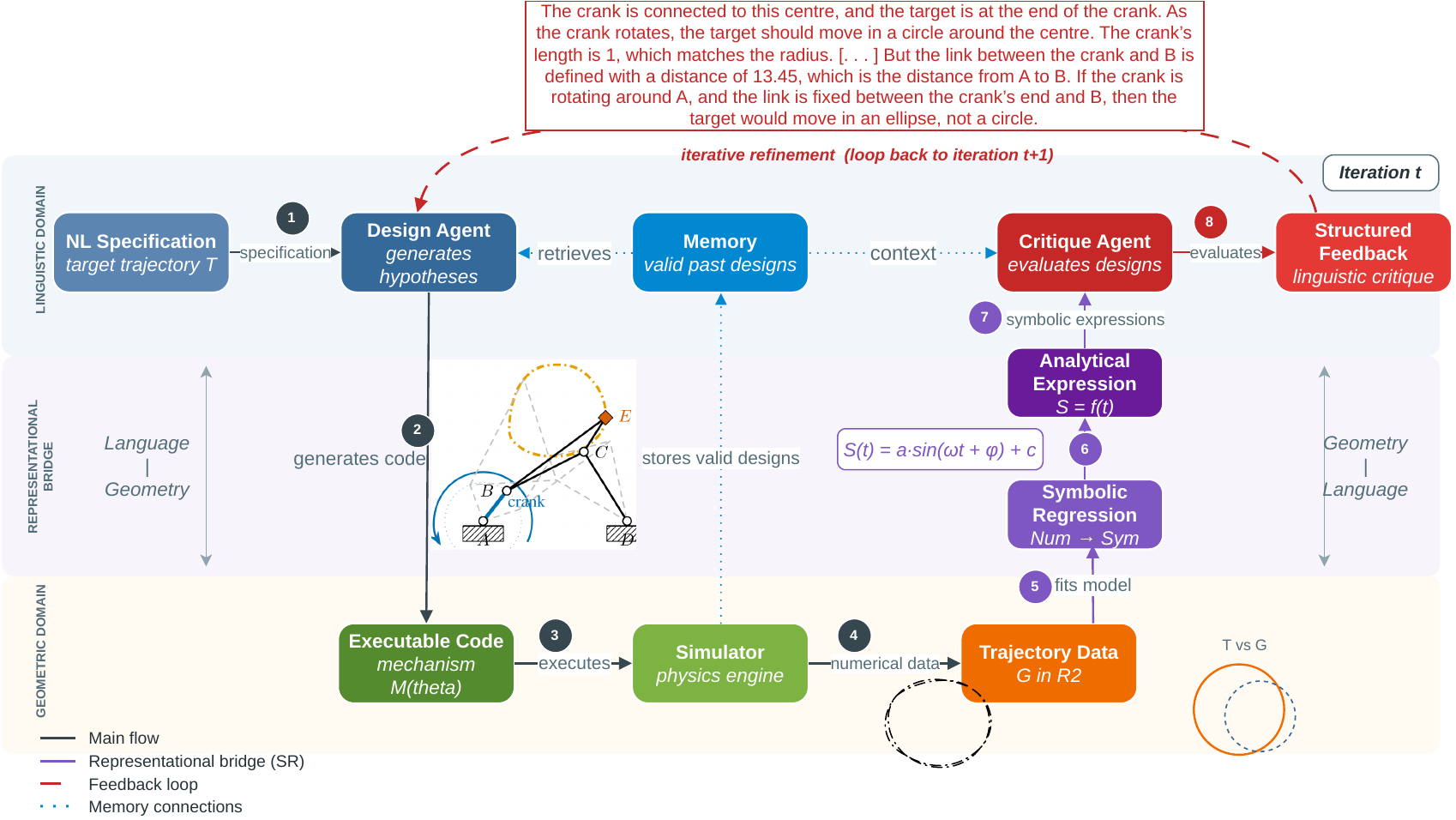}
    \caption{The linguistic--numerical interface in three layers. In the \emph{linguistic domain} a design agent proposes mechanism hypotheses and a critique agent evaluates them; in the \emph{geometric domain} executable code drives a simulator to yield a trajectory; the \emph{representational bridge} converts that trajectory, via symbolic regression, into an analytical expression the agents can read. Numbered arrows (1--8) trace the closed loop from the natural-language target back to the design agent. A persistent memory of valid designs feeds both agents.}
    \label{fig:method}
\end{figure*}

Formally, let $\pi_\theta$ denote a pre-trained language model; we iteratively sample mechanism designs $\mathcal{F} = \{\mathcal{M}ech : \mathcal{M}ech \sim \pi_\theta\}$, aiming to minimise $\mathcal{M}ech^* = \arg\min_{\mathcal{M} \in \mathcal{M}ech} d(\mathcal{T},\mathcal{G}_\mathcal{M})$. The method couples a \acf{da} (linguistic side) with a \acf{ca} (reasoning over symbolic translations of the simulator's output) in a closed cycle: natural language $\to$ executable code $\to$ geometric simulation $\to$ \textbf{symbolic interface} $\to$ linguistic critique $\to$ refined design (Fig.~\ref{fig:method}; full procedure in Algorithm~\ref{alg:method}).

\subsection{Design Agent: From Language to Geometry}
\label{sec:mda}

The \ac{da} translates a linguistic specification into a geometric design hypothesis. Given a mechanism specification $\mathcal{M}(\theta)$, Simulator $\mathcal{S}im$, Set of examples $\mathcal{E}x$, Target points along the end effector trajectory $\mathcal{P}$, and Analytical equation of the target path $\mathcal{T}$, the \ac{da} applies an evaluation function:
$$\mathbb{D}_a\left(\mathcal{M}(\theta), \mathcal{S}im, \mathcal{E}x, \mathcal{P}, \mathcal{T} \right) \rightarrow \mathbb{M}ech$$
\noindent Here the \ac{da} moves from a natural-language task description (prompt and symbolic specifications) to a mechanical hypothesis $\mathbb{M}ech$, an instantiation under simulator $\mathcal{S}im$ satisfying $\mathcal{M}(\theta)$, targeting a geometric trajectory in Cartesian ($\mathbb{R}^2$) space. At each iteration $t$, we sample a batch of $b$ mechanism configurations
$F_t = \{f_i\}_{i=1}^b,\quad f_i \sim \mathbb{D}_a(\cdot \mid p_t),$
where $p_t$ is the constructed prompt (App.~\ref{sec:dap}). Each candidate is validated by simulation in $\mathcal{S}im$; infeasible ones are discarded to prevent ill-defined features from propagating to later iterations.
Thus, the \ac{da} goes from: linguistic prompt $\to$ mechanical candidate $\to$ geometric/\ac{ca} validation $\to$ refined prompt.

The prompt is structured in four abstraction stages (motion-profile alignment with $\mathcal{T}$, structural-constraint enforcement, iterative refinement via feedback, and sequential code generation; see App.~\ref{sec:cap}), composed as $\mathcal{P}_d = f_{\text{M}em} \circ f_{\text{C}on} \circ f_{\text{E}x} \circ f_{\text{S}im} (\text{Input})$, where $f_{\text{S}im}$ embeds simulator/mechanism documentation, $f_{\text{E}x}$ supplies examples, $f_{\text{M}em}$ injects accumulated designs, and $f_{\text{C}on}$ encodes key points and target equations. The \ac{da} refines parameters $\theta$ via feedback-driven updates and leverages \ac{sr} \cite{doi:10.1126/science.1165893, petersen2021deep, shojaee2025llmsr, shojaee2023transformerbased, meidani2024snip, udrescu2020aifeynman, lacava2021srbench, kamienny2022endtoend} to extract expressions that guide optimisation.


\subsection{Critique Agent: From Geometry Back to Language}
\label{sec:mca}
The critique agent performs the inverse representational transformation, mapping the simulator's geometric output back into structured linguistic analysis that guides the next iteration. Given mechanism $\mathcal{M}(\theta)$, simulator output $\mathcal{S}$, memory $\mathcal{M}em$, and designer response $\mathcal{R}$, it applies $\mathbb{C}_a\left(\mathcal{M}(\theta), \mathcal{S}, \mathcal{M}em, \mathcal{R} \right) \rightarrow \mathbb{F}$, producing feedback $\mathbb{F}$ informed by mechanical-design principles, kinematic performance analysis, and historical context (App.~\ref{sec:cap}). The \ac{ca} evaluates four criteria, (1) \textbf{correctness}, (2) \textbf{error identification}, (3) \textbf{complexity}, (4) \textbf{refinement suggestions}, through the prompt composition $\mathcal{P}_c = f_{\text{mem}} \circ f_{\text{sim}} \circ f_{\text{ex}} \circ f_{\text{des}} (I)$, where $f_{\text{des}}$ ingests the \ac{da}'s output, $f_{\text{sim}}$ aligns with fresh simulation/\ac{sr} data, $f_{\text{ex}}$ standardises critique style, and $f_{\text{mem}}$ grounds the evaluation in past designs.


\label{sec:sr}
\paragraph{Symbolic regression as the interface layer.} The central technical insight of our approach is that converting a numerical trajectory into a compact symbolic form creates a structured interface between the simulator's numerical outputs and the \ac{llm}'s linguistic reasoning. \ac{sr} transforms raw simulation traces into concise analytical expressions that expose functional structure (periodicity, symmetry, monotonic segments) in a form that language models can directly interpret and reason about.

Let \(\mathcal{G}\) be the simulator trace and \(S\) a symbolic surrogate. We fit \ac{sr} by minimising:
$
\mathcal{L}(S) = d\big(\mathcal{G}, \mathcal{T}\big)
$, where $d(\cdot,\cdot)$ is the trajectory error. Both the design agent and critique agent operate on this shared symbolic vocabulary, enabling principled comparison between target and generated trajectories, identification of specific geometric deficiencies (e.g., missing periodicity, incorrect curvature), and transparent communication of reasoning traces to human reviewers.


\subsection{Revision: Closing the Representational Loop}
\label{sec:mrev}
The revision function integrates linguistic feedback from the critique agent with geometric evidence from the simulator to produce a refined design: $\mathbb{R}_p\left(\mathcal{D}_r, \mathcal{C}_r, \mathcal{S}, \mathcal{M}em \right) \rightarrow \mathcal{D}'_r$, where $\mathcal{D}_r$ is the current designer response, $\mathcal{C}_r$ the critique response, $\mathcal{S}$ the simulator output, and $\mathcal{M}em$ the memory repository. At iteration $t$, $\mathcal{D}'_{r,t} = \mathbb{R}_p(\mathcal{D}_{r,t-1}, \mathcal{C}_{r,t-1}, \mathcal{S}_{t-1}, \mathcal{M}em_{t-1})$, subject to three constraints: \textbf{(i) feedback adherence}, $\forall f \in \mathcal{C}_{r,t-1}\;\exists m \in \mathcal{D}'_{r,t}:\mathsf{Addresses}(m,f)=1$; \textbf{(ii) simulation consistency}, $\mathcal{V}(\mathcal{M}(\theta)'_{r,t},\mathcal{S}_{t-1})=1$; and \textbf{(iii) efficiency}, $\mathsf{Complexity}(\mathcal{M}(\theta)'_{r,t}) \le \mathsf{Complexity}(\mathcal{M}(\theta)_{r,t-1})$ subject to $d(\mathcal{D}'_{r,t},\mathcal{T}) \le d(\mathcal{D}_{r,t-1},\mathcal{T})$.

\paragraph{Learning through iterative refinement.} Early iterations address coarse geometric misalignment (wrong trajectory shape); later iterations target curvature matching and parameter tuning. This adaptive progression emerges from critique feedback: when $d(\mathcal{D}_{r,t},\mathcal{T})$ is large the \ac{ca} flags structural issues, and as $d$ decreases critiques shift to fine parametric adjustments. The revision prompt $\mathcal{P}_r = f_{\text{mem}} \circ f_{\text{sim}} \circ f_{\text{crit}} \circ f_{\text{des}} (\text{Input})$ (App.~\ref{sec:darp}) layers designer response, critique, simulation/\ac{sr} evidence, and memory. The loop terminates when $d(\mathcal{M}(\theta)'_{r,t}, \mathcal{T}) \le \epsilon$ or $t \ge R_{max}$.

\paragraph{Memory.} A persistent repository $M_t = \{m_1,\dots,m_n\}$ stores only mechanisms that pass the validity check $\mathcal{V}(m_i)=1$; retrieval selects the top-$k$ designs by $\arg\min_{m_i \in M_t} d(m_i, m_{\text{target}})$. This selective storage preserves validity, eliminates known-invalid designs from future context, and biases sampling toward geometric patterns proven for similar trajectories. As iterations proceed, the memory acts as a sufficient statistic for task-relevant knowledge, yielding lower expected error with fewer iterations as $|M_t|$ grows (full learning-dynamics analysis in App.~\ref{app:memory_dynamics_method}).

\section{Empirical Analysis}
\label{sec:results}

\noindent \textbf{Summary of findings.} Across three model families and a full-factorial ablation, we present the first evidence that an \ac{sr}-derived symbolic intermediary acts as a learning interface between an \ac{llm} and a physics simulator, letting language models \emph{learn} continuous geometric design from natural language without any parameter updates. The four research questions below show, in turn, that geometric error decreases over iterations, that critique feedback is the dominant gain channel, that the symbolic layer grounds critique in geometric verification, and that at a matched budget the interface outperforms a genetic-algorithm baseline by 19--53\%.

\noindent \textbf{Experimental Setup.} We evaluate on a synthetic dataset of analytically defined trajectories covering six analytic planar curves of increasing geometric complexity: circles, ellipses, lines, parabolas, \ac{lb}, and NACA four-digit airfoils \cite{jacobs1933naca460,abbott2012theory}; mean Chamfer distance orders them from NACA airfoils (easiest) through ellipses, lines, circles, and \ac{lb} to parabolas (hardest). These curves are analytically defined, which supplies exact ground truth for the symbolic interface, and together they span distinct curve families (linear, conic, transcendental, and engineered airfoil profiles); the resulting difficulty is kinematic rather than algebraic, since the open parabola is hardest despite its low degree (full selection criteria in App.~\ref{app:shape_eligibility}). For each shape class we sample 5 random instances and draw $n=4$ target points along each curve (App.~\ref{app:dataset_details}, \ref{sec:shape_complex}). At each iteration the design agent generates 3 candidate mechanisms (sampling temperature $=0.8$); we set $R_{\max}=20$ and convergence threshold $\epsilon=0.05$. As a baseline we use a two-stage genetic algorithm (GA) that exhaustively enumerates kinematic topologies and then optimises dimensional parameters; Enum+GA is the established, budget-controllable comparator in computational kinematics, and we justify this choice in App.~\ref{app:baseline_eligibility}. Simulation uses PyLinkage \cite{Farajallah2024pylinkage} and symbolic regression uses PySR \cite{cranmer2023interpretablemachinelearningscience}; the iterative closest point algorithm \cite{richardos_icp, lu1997robot} aligns simulated and target point clouds before evaluation. Geometric fidelity is the Chamfer distance ($d_{Chamfer}$) between generated and target trajectories (after ICP alignment); \textit{execution success} is a binary indicator of whether a generated mechanism compiles and runs without errors (App.~\ref{app:metrics}); we also report iteration count ($t_{iter}$).

\noindent \textbf{Ablation design.} We ran a full-factorial experiment varying six factors: (i) \textbf{Model} $\in\{\text{\ac{llama}},\ \text{\ac{gemma}},\ \text{\ac{qwen}}\}$; (ii) \textbf{Shape} (all target trajectories); (iii) number of retrieved design examples (\#Ex $\in\{2,3\}$); (iv) \textbf{Memory} $\in\{0,2\}$; (v) \textbf{Critique feedback} $\in\{\text{No},\text{Yes}\}$; and (vi) \textbf{Symbolic feedback (SFB)} $\in\{\text{No},\text{Yes}\}$. Every combination was executed; the full results table is in App.~\ref{sec:all_quant}.

\subsection{Results}

We organise the evidence around four research subquestions, each tied to a stated contribution (\S\ref{sec:intro}) and the corresponding finding in the Discussion (\S\ref{sec:discussion}).

\paragraph{RQ1: To what extent does the interface produce learning without parameter updates?}
\label{rq:1}
\emph{(Substantiates Contribution~(1); supports \S\ref{sec:discussion}, ``Learning without gradient updates''.)} \textbf{The method learns geometric design without any parameter update, with error falling as the agentic design, critique, and revision loop accumulates feedback.} We adopt Mitchell's operational definition: a program learns from experience $E$ at a task $T$ under measure $P$ if its $P$ on $T$ improves with $E$. Here $T$ is planar mechanism synthesis, $P$ is the Chamfer distance, and $E$ is the accumulated simulation feedback, symbolic trajectory analysis, and memory, no parameter is ever updated. By this criterion the method demonstrably learns. The representational cycle (Sections~\ref{sec:mda}--\ref{sec:mrev}) produces convergent trajectories across all three model families: even \emph{without} critique feedback the first iterations already cut geometric error substantially (up to 77.1\% reduction), and adding critique-driven feedback compounds this, with per-shape Chamfer distance falling by up to 90\% (e.g.\ \ac{llama} on circles, $12.239\!\to\!1.466$). The decisive evidence is the \emph{shape} of the trajectory rather than any single number: error decreases as experience accumulates, the operational signature of learning. The two remaining hallmarks of this paradigm are quantified in the appendix: \emph{knowledge transfer} via memory (App.~\ref{app:memory_dynamics}; e.g.\ \ac{qwen} aggregate mean falls by $\approx\!11.6$ with memory), \emph{failure avoidance} via selective storage of executable designs (App.~\ref{app:memory_dynamics_method}), and the \emph{emergent curriculum}, structural commentary declining significantly over iterations while parametric refinement persists ($\chi^2\!=\!278.67$, $p\!\approx\!1.5\!\times\!10^{-62}$; App.~\ref{app:curriculum}).

\begin{figure}[t]
\includegraphics[width=0.9\columnwidth]{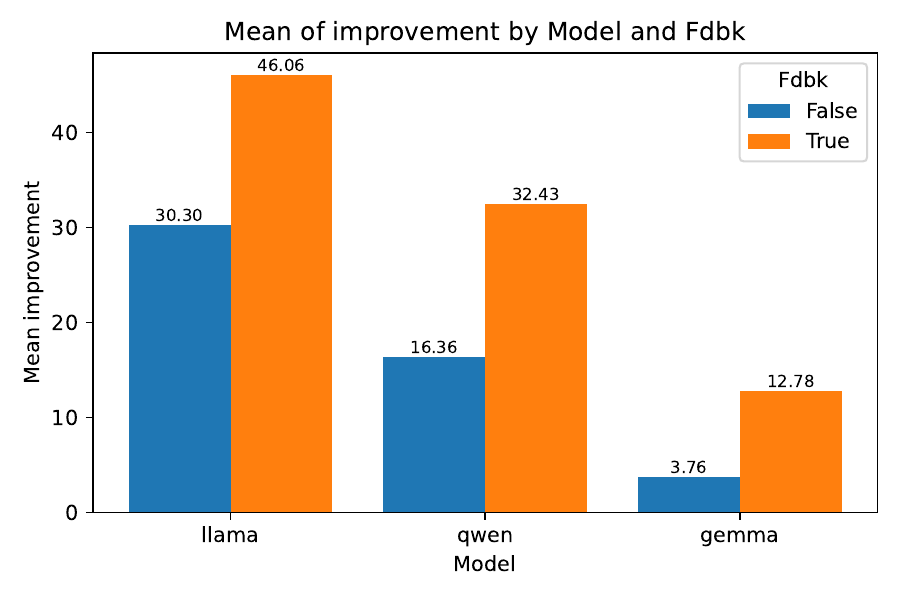}
  \caption{Average percentage improvement in Chamfer distance with feedback (Fdbk) vs.\ without; the largest uplift is \acl{llama} (30\%$\to$46\%), consistent across all three families.}
  \label{fig:agent_feedback}
\end{figure}

\paragraph{RQ2: Which interface components drive the gain, and to what degree?}
\label{rq:2}
\emph{(Substantiates Contributions~(1) and~(3); supports \S\ref{sec:discussion} on critique vs.\ symbolic layer.)} \textbf{Critique feedback carries most of the gain, and symbolic regression adds a consistent increment on the hardest curves.} Critique-based feedback contributes the largest gain, consistently across model families. Semantic success improves for all three models when feedback is enabled, with relative gains of up to roughly 88\%. Execution validity moves in lockstep, rising by 9 to 24 percentage points depending on the model. Critically, feedback both \emph{increases} refinement rounds (one to four on average) and \emph{lowers} Chamfer distance, so the extra rounds improve quality rather than prolong search; a paired Wilcoxon signed-rank test confirms the reduction is significant. Figure~\ref{fig:agent_feedback} shows the corresponding Chamfer gains across all three families, with the largest uplift for \acl{llama} (30\%$\to$46\%). The symbolic-regression layer then adds capability \emph{on top of} critique, and the gain is a consistent trend rather than a per-model artefact: in Table~\ref{tab:model_vs_ga_main}, for every model and on both the 4-bar and 6-bar topologies the SR+Feedback condition yields a lower normalised Chamfer than the No-Feedback/No-SR baseline. Globally, experiments with SR+Feedback reach a mean outcome of 140.00 vs.\ 148.50 without SR ($\Delta\!\approx\!8.5$), and restricting to $Mem\!=\!2$ preserves the gap ($\Delta\!\approx\!6.94$). The reductions concentrate on the hardest curves (parabola, $\Delta\!\approx\!33.78$; ellipse, $0.31$; \ac{lb}, $0.27$; NACA, $0.11$), indicating that the symbolic layer pays off most \emph{where} multi-step analytical decomposition is required.

\paragraph{RQ3: In what ways does the interface change the \emph{nature} of reasoning, beyond the score?}
\label{rq:3}
\emph{(Substantiates Contribution~(2); supports \S\ref{sec:discussion}, ``a new form of geometric reasoning''.)} \textbf{The symbolic interface grounds reasoning, turning qualitative structural commentary into quantitative geometric verification.} Paired critique transcripts (Fig.~\ref{fig:qualitative_comparison_app}; full SR-vs-no-SR transcripts in App.~\ref{app:critique_comparison}) show the symbolic layer shifts reasoning from qualitative structural commentary to quantitative geometric verification. On the parabola, the canonical case, the SR-enabled critique identifies the curve class from $y=(x-7)^2+3$ and applies the law of cosines ($40^2=30^2+20^2-2\cdot 30\cdot 20\cos\theta\Rightarrow\theta\!\approx\!104.5^\circ$) to check joint feasibility, whereas the matched SR-absent critique only states that ``a linear slider can only move along a straight path''; this ties the qualitative pattern directly to the largest quantitative gain ($\Delta\!\approx\!33.78$). The same substitution of an explicit numerical check for a vague structural assertion recurs across shapes: on a NACA airfoil the SR-enabled critique computes explicit residuals ($\sqrt{2.25+1}\!\approx\!1.8$ vs.\ target $2.5$) while the SR-absent run defaults to ``the four-bar linkage is a common choice'', and on a circle it diagnoses an ellipse-producing parameter by explicit calculation rather than generic commentary. Programmatically scoring all critique entries with an eight-category taxonomy quantifies this: engagement with discovered symbolic structure rises with \ac{sr} for \emph{every} model. The shift is one of \emph{grounding} rather than raw frequency, and is architecture-dependent, the reasoning-trained \ac{qwen}~4B is where \ac{sr} most increases grounded analytical reasoning; an aggregate-vs-per-model reconciliation (the effect is a Simpson's paradox at the pool level) is given in App.~\ref{app:reasoning_taxonomy}.

\paragraph{RQ4: By how much does it outperform numerical search at matched budget, and how does model capacity interact with the bridge?}
\label{rq:4}
\emph{(Substantiates Contribution~(3); supports \S\ref{sec:discussion}, ``Why architecture matters''.)} \textbf{The method outperforms the GA by 19--53\% at a matched budget, and capacity interacts non-monotonically with the bridge.} Table~\ref{tab:model_vs_ga_main} reports normalised Chamfer distances relative to the budget-matched GA baseline (Pop:\,6 / Gen:\,20; 120 fitness evaluations) for both 4-bar and 6-bar topologies. Two findings emerge. \textit{First}, critique-driven refinement alone achieves significant improvements over the GA on 4-bar mechanisms and, at comparable budget, on 6-bar mechanisms as well: linguistic reasoning with simulation feedback outperforms population-based numerical search. \textit{Second}, adding the symbolic-regression layer (FB + SR) broadens this advantage: \ac{llama} and \ac{qwen} achieve 31--36\% improvement on both topologies, with \ac{qwen} more than doubling its margin from FB-only to FB+SR on 4-bar mechanisms (15.5\%$\to$31.2\%). The interaction with capacity is not monotone in scale: the compact reasoning-trained \ac{qwen}~4B outperforms the $17\times$ larger dense \ac{llama}~70B on the hardest curve (parabola), while the ordering reverses on simpler shapes, indicating that the bridge pays off most when the base model's training regime supplies the right inductive bias for multi-step analytical decomposition (per-shape numbers in App.~\ref{sec:all_quant}; mechanism discussed in \S\ref{sec:discussion}).

\begin{table}[t]
\centering
\small
\setlength{\tabcolsep}{3pt}
\renewcommand{\arraystretch}{1.15}
\begin{tabular}{@{} l l r r @{}}
\toprule
\textbf{Model} & \textbf{Condition} & \textbf{Value} & \textbf{Imp.\,\%} \\
\midrule
\multicolumn{4}{c}{\textit{4-bar (matched budget: Pop:\,6 / Gen:\,20)}} \\
\midrule
Gemma 12B & No FB, No SR & 1.319 & \textcolor{red}{$-$31.9\%\,$\downarrow$} \\
          & FB           & 1.219 & \textcolor{red}{$-$21.9\%\,$\downarrow$} \\
          & FB + SR      & 1.185 & \textcolor{red}{$-$18.5\%\,$\downarrow$} \\
\cmidrule(l){1-4}
Llama 70B & No FB, No SR & 0.742 & \textcolor{teal}{$+$25.8\%\,$\uparrow$} \\
          & FB           & $\star$\,\textbf{0.634} & \textcolor{teal}{$\boldsymbol{+}$\textbf{36.6\%}\,$\uparrow$} \\
          & FB + SR      & 0.659 & \textcolor{teal}{$+$34.1\%\,$\uparrow$} \\
\cmidrule(l){1-4}
Qwen 4B   & No FB, No SR & 1.107 & \textcolor{red}{$-$10.7\%\,$\downarrow$} \\
          & FB           & 0.845 & \textcolor{teal}{$+$15.5\%\,$\uparrow$} \\
          & FB + SR      & 0.688 & \textcolor{teal}{$+$31.2\%\,$\uparrow$} \\
\midrule
\multicolumn{4}{c}{\textit{6-bar (matched budget: Pop:\,6 / Gen:\,20)}} \\
\midrule
Gemma 12B & No FB, No SR & 1.232 & \textcolor{red}{$-$23.2\%\,$\downarrow$} \\
          & FB           & 1.154 & \textcolor{red}{$-$15.4\%\,$\downarrow$} \\
          & FB + SR      & 1.112 & \textcolor{red}{$-$11.2\%\,$\downarrow$} \\
\cmidrule(l){1-4}
Llama 70B & No FB, No SR & 0.712 & \textcolor{teal}{$+$28.8\%\,$\uparrow$} \\
          & FB           & $\star$\,\textbf{0.634} & \textcolor{teal}{$\boldsymbol{+}$\textbf{36.6\%}\,$\uparrow$} \\
          & FB + SR      & 0.648 & \textcolor{teal}{$+$35.2\%\,$\uparrow$} \\
\cmidrule(l){1-4}
Qwen 4B   & No FB, No SR & 1.063 & \textcolor{red}{$-$6.3\%\,$\downarrow$}  \\
          & FB           & 0.829 & \textcolor{teal}{$+$17.1\%\,$\uparrow$} \\
          & FB + SR      & 0.698 & \textcolor{teal}{$+$30.2\%\,$\uparrow$} \\
\bottomrule
\end{tabular}
\caption{Normalised Chamfer distance vs.\ the budget-matched GA baseline for 4-bar and 6-bar mechanisms (120 fitness evaluations). \textbf{Value}~$=$ mean over shapes of $\tfrac{\mathrm{LLM\;median}}{\mathrm{GA\;median}}$; values $<1$ indicate the LLM outperforms the GA. Arrows on Imp.\,\% mark direction; $\star$ marks the best result per topology. \textcolor{teal}{Teal} = improvement; \textcolor{red}{red} = degradation. High-budget (Pop:\,60 / Gen:\,300) and low-budget (Pop:\,3) results are in Appendix~\ref{sec:all_quant}.}
\label{tab:model_vs_ga_main}
\end{table}

\section{Discussion}
\label{sec:discussion}

\paragraph{The symbolic interface enables a new form of geometric reasoning.}
Beyond data compression, the symbolic-regression interface changes \emph{what} the model reasons over: instead of opaque coordinate arrays, the critique agent receives an explicit functional form it can inspect, compare against the target, and turn into specific parameter recommendations; without it the agent sees only scalar distances and produces generic, less actionable feedback. The empirical signature of this shift, quantitative geometric verification replacing qualitative structural commentary, is established in RQ3 (\S\ref{rq:3}); the effect is architecture-dependent and, in aggregate, masked by a compositional (Simpson's) artefact reconciled in App.~\ref{app:reasoning_taxonomy}.

\paragraph{Learning without gradient updates: characterising the paradigm.}
RQ1 (\S\ref{rq:1}) evidences three hallmarks of learning: progressive improvement, knowledge transfer via memory, and failure avoidance through selective storage of executable designs. Under Mitchell's criterion (task $=$ synthesis, performance $=$ Chamfer distance, experience $=$ feedback and memory), this constitutes learning without any parameter update.

This learning paradigm occupies a distinctive position among machine learning approaches. It differs from standard \emph{in-context learning} \cite{brown2020language,min2022rethinking,garg2022incontext}, whose examples here are generated dynamically through simulation interaction rather than fixed, each iteration adding design--outcome pairs that reshape the context; from \emph{meta-learning} \cite{finn2017model}, with no outer-loop optimisation over task distributions, adaptation coming instead from accumulated memory and critique; and from \emph{heuristic search}, whose operators here are not fixed, as the model adapts its proposals from symbolic analysis of past failures. We term this \emph{interface-mediated in-context learning}: the interface channels the model's pre-trained reasoning toward domain-specific objectives through structured feedback and dynamically accumulated experience.

The critique-driven refinement process also exhibits an emergent curriculum: large geometric error elicits structural critiques (wrong topology, disconnected links) while small error elicits fine-grained parametric adjustments (link-length ratios, joint angles). This progression arises naturally from the interaction between the distance metric and the symbolic representation, without explicit curriculum design, and is quantified in RQ1 (\S\ref{rq:1}; structural commentary declines significantly over iterations, $\chi^2$ test in App.~\ref{app:curriculum}).

\paragraph{Why architecture matters: inductive biases for geometric reasoning.}
Our results (RQ4, \S\ref{rq:4}) reveal a non-trivial interaction between model architecture and the capacity to exploit symbolic representations, with the model ordering reversing between simpler shapes and the hardest curves. Step-by-step reasoning training, a hallmark of \ac{lrm}-style models, provides inductive biases particularly valuable when the geometric problem requires multi-step analytical decomposition (as parabolas do), whereas the broader pre-trained knowledge of larger dense models better supports shapes where direct pattern matching suffices \cite{wei2022cot,zelikman2022star}. The same pattern holds under symbolic feedback (\ac{qwen} shows the largest absolute improvements \cite{lightman2024verify}) and memory-based transfer, where compact reasoning-trained models benefit consistently while very large dense models can suffer interference at higher example counts (App.~\ref{app:memory_dynamics}). The choice of base model is therefore governed by whether its training regime aligns with the reasoning demands of the representational bridge, rather than by scale alone.

The interface principle could extend beyond mechanism synthesis to other domains that satisfy simulability, symbolic approximability, and iterative refinability, such as molecular geometry optimisation, robotic trajectory planning, structural topology optimisation, and metamaterial design (App.~\ref{app:generalisability}). Methodologically, the paradigm extends tool-augmented LMs, neuro-symbolic integration, and retrieval-augmented generation by using discovered (rather than hard-coded) symbolic content and dynamically accumulated (rather than static) retrieval \cite{lewis2020rag} (App.~\ref{app:broader_connections}).

\section{Conclusion}
\label{sec:conclusion}

We presented a self-guided design, critique, and revision loop in which language models and physics simulators coordinate through symbolic intermediaries: compact analytical expressions discovered via symbolic regression translate continuous simulator outputs into a form that \ac{llm}s can interpret, compare, and critique, turning a sequence of isolated proposals into iterative, memory-driven refinement without modifying the model. On planar mechanism synthesis it is both effective and convergent: critique feedback through the interface is the most transformative component (up to 90\% improvement in geometric fidelity; \S\ref{rq:1},~\ref{rq:2}), and the symbolic layer adds further capability, most pronounced in models with appropriate inductive biases (\S\ref{rq:4}). The method thus learns through memory-based transfer and iterative refinement without any parameter updates, a principle that extends to any domain where simulator-produced numerical behaviour must be interpreted, critiqued, and acted upon linguistically.

\section{Limitations}
\label{sec:limitations}

Two aspects of our setup delimit the present scope and point to natural extensions. First, we deliberately evaluate on 2D planar curves, a controlled setting that isolates the reasoning loop; because the symbolic interface is agnostic to trajectory dimensionality, extending it to 3D paths, time-varying fields, and richer geometric constraints is a natural next step. Second, the method presently takes analytical target equations as input, a reasonable assumption in many engineering settings, and the interface is in principle compatible with softer specifications such as hand-drawn sketches or verbal descriptions, a promising avenue for broadening applicability.

\section{Ethical Statement}

This work explores AI-driven methods for automating engineering design tasks. While the results are promising, the proposed method is exploratory and should not replace expert engineering judgement.


\bibliography{latex/custom}

\appendix
\label{sec:appendix}
\section{Appendix: Example}

Figure~\ref{fig:example} shows a materialisation of the neuro-symbolic synthesis method for planar mechanism design. \textbf{(1)} \acl{da} generate hypotheses, \textbf{(2)} simulator simulates and \acl{ca} critiques and prescribe revisions, and \textbf{(3)} \acl{da} performs the revision then repeat.

\begin{figure*}
    \centering
    \includegraphics[width=1.0\linewidth,height=1.0\textheight]{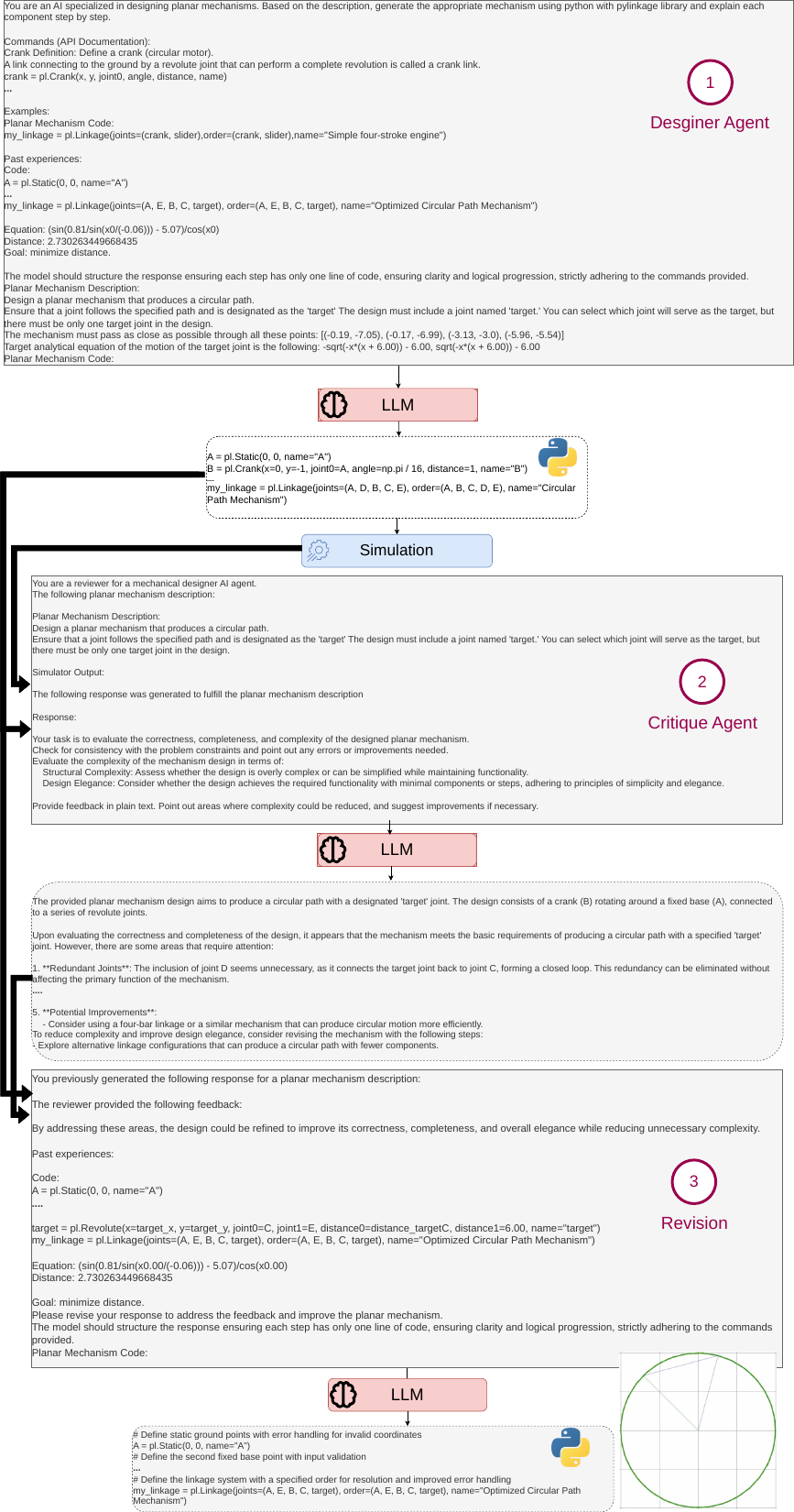}
\end{figure*}
\begin{figure*} [t!]
    \caption{Neuro-symbolic closed-loop for planar mechanism synthesis: (1) \acl{da} generates executable mechanism hypotheses from the specification and memory; (2) Executor runs candidates and \acl{ca} critiques and issues structured revision directives; (3) \acl{da} performs the revision, then repeats. The circle in the picture represents the path traced by the mechanism as simulated by the simulator.}
    \label{fig:example}

\end{figure*}

\paragraph{1. Designer Agent.}
This agent takes a target specification, either an analytic curve or sampled points, together with high-level constraints such as naming conventions, plus memory of previously validated designs, and produces executable mechanism hypotheses. It translates abstract requirements into concrete structural proposals by exploring topology and parameter spaces while enforcing hard constraints. The output is an executable representation of a mechanism (topology and parameter values) expressed in the system's canonical design language. Emphasis is on generating a diverse set of simulation-ready candidates, biasing toward promising motifs learned from memory and avoiding known dead-ends.

\paragraph{2. Simulation and Critique.}
Each candidate is validated in the simulation to confirm physical executability and to convert numeric behaviour into compact, interpretable summaries. The simulation produces time-series traces of the end-effector in $R^2$, execution diagnostics (for example, success or failure), and a symbolic surrogate, a compact analytic expression that approximates the numeric trace. This stage gates validity by rejecting non-executable or unstable designs, and it exposes algebraic and functional motifs in the motion through the symbolic summary.

\paragraph{3. Revision.}
The Revision step consumes the critique entries into structured, actionable modifications. It produces a new revised synthesises hypothesis that implements those changes. The goal is to recommend minimal, high-impact edits that improve fit and robustness while reducing unnecessary complexity, using memory to prevent repeating known failures and to suggest effective design motifs.

\begin{enumerate}
  \item The Designer proposes one or more simulation-ready mechanism hypotheses.
  \item The Simulator simulates each candidate and produces numeric traces, diagnostics, and symbolic surrogates.
  \item The Critic analyses.
  \item The Designer ingests the directives and memory items, revises or generates new hypotheses, and the cycle repeats.
\end{enumerate}

\section{Appendix: Method Algorithm}

In this section, we provide a detailed description of our dual-agent iterative design loop for synthesising planar mechanisms (see Algorithm~\ref{alg:method}). Starting with an initial set of parameters, the \ac{da} proposes a candidate mechanism, which is then simulated on $\mathcal{S}im$. Successful simulations trigger a distance-metric evaluation against the target trajectory $\mathcal{T}$, and \ac{sr} feedback. The validated design is then stored in memory. The \ac{ca} then assesses performance and provides feedback on how to refine the design, prompting the \ac{da} to generate updated parameters for a second simulation and re-evaluation. This cycle of design, simulation, evaluation and refinement repeats until the mechanism meets the predefined convergence or performance criteria.

\newpage

\begin{algorithm}[H]
\caption{Iterative Planar Mechanism Design with Dual Agent, Simulation, and Evaluation}
    \begin{algorithmic}[1]
    \State \textbf{Input:} Initial design parameters, Simulator on \(\mathcal{S}im\), evaluation functions, target trajectory \(\mathcal{T}\)
    \State \textbf{Output:} Optimized planar mechanism design \(\mathcal{M}(\theta)\)
    \vspace{0.5em}
    \While{Design objective not met}
        \State \textbf{Design:} Generate a new planar mechanism \(\mathcal{M}(\theta)\) using the \ac{da}.
        \State \textbf{First Simulation:} Simulate \(\mathcal{M}(\theta)\) on \(\mathcal{S}im\)
        \If{Simulation successful}
            \State Compute distance metric \(d(\mathcal{M}(\theta), \mathcal{T})\)
            \State Discover \ac{sr} equation
            \State Store \(\mathcal{M}(\theta)\) in memory
        \EndIf
        \begin{itemize}
            \item \ac{ca} evaluates the mechanism and provides feedback
        \end{itemize}
        \State \textbf{Refinement:} Update the design strategy using feedback from the evaluation
        \State \textbf{Design (Refinement):} Generate a refined mechanism \(\mathcal{M}^\prime(\theta)\) using the updated strategy.
        \State \textbf{Simulation:} Simulate the refined mechanism \(\mathcal{M}^\prime(\theta)\) on \(\mathcal{S}im\)
        \If{Simulation successful}
            \State Store \(\mathcal{M}^\prime(\theta)\) in memory
        \Else
            \State Discard \(\mathcal{M}^\prime(\theta)\)
            \State \textbf{continue} to the next iteration
        \EndIf
        \State \textbf{Evaluation:} 
        \begin{itemize}
            \item Re-compute the distance metric \(d(\mathcal{M}^\prime(\theta), \mathcal{T})\)
            \item Re-assess \ac{sr} feedback
        \end{itemize}
    \EndWhile
    \State \Return Best mechanism \(\mathcal{M}(\theta)\) or refined mechanism \(\mathcal{M}^\prime(\theta)\) meeting design criteria
    \end{algorithmic}
    \label{alg:method}
\end{algorithm}

\clearpage

\section{Appendix: Evaluation Metrics}
\label{app:metrics}

\noindent \textbf{Point Set Alignment.} For a generated mechanism $m_t$ that yields an end effector trajectory $GeneratedPoints$, we first perform point set registration via the \ac{icp} algorithm to establish the correspondence with the target motion profile. The algorithm aligns the generated point set with the target point set through rigid transformations, which compensates for the differences in the orientation of the coordinate systems.

\noindent \textbf{Chamfer distance} The Chamfer distance between two point sets \(P\) and \(Q\) measures their similarity by summing, for each \(p\in P\), the distance to its nearest neighbour in \(Q\), and doing the same from \(Q\) to \(P\). After trajectory alignment, the mechanism's performance is quantified using the Chamfer distance, which provides this bidirectional measure of spatial proximity between the two point sets:
Formally, one common formulation is \ref{fig:chamfer}.

\begin{figure*}[htbp!]
\centering
\begin{align*}
  d_{Chamfer}(P, Q) = \frac{\sum_{p \in P} \min_{q \in Q} \|p - q\|^2}{|P|} + \frac{\sum_{q \in Q} \min_{p \in P} \|q - p\|^2}{|Q|}
\end{align*}
\caption*{}
\phantomcaption
\label{fig:chamfer}
\end{figure*}

\noindent \textbf{Semantic Success Rate} We define \emph{Semantic success} as a binary indicator of whether a generated program yields an executable. Formally, given a program \(P\), we denote its execution in the simulator as \(T_{\text{pred}} = \text{Sim}(P)\). Then
\begin{equation}
\resizebox{\columnwidth}{!}{$
S(P) =
\begin{cases}
1, & \text{if } P \text{ parses and executes without error},\\
0, & \text{otherwise.}
\end{cases}
$}
\end{equation}

\noindent Implementation notes: any parse, compilation, runtime error, timeout, or early termination that prevents a simulation from being produced is treated as a failure (i.e., \(S=0\)). All Python code generated by the \ac{llm} was automatically formatted with \texttt{black} to ensure a consistent style across implementations. Using \texttt{black} produces deterministic, idempotent formatting that reduces stylistic variation introduced by code generation and simplifies code inspection and review.

\section{Appendix: Dataset Details}
\label{app:dataset_details}
Let
\begin{align*}
  \mathcal{S}
  = \bigl\{\,
    \mathcal{C}(r,x_1,y_1),\; \\
    \mathcal{E}(a,b,x_1,y_1),\; \\
    \mathcal{L}(x_1,y_1,x_2,y_2),\; \\
    \mathcal{P}(a,h,k),\; \\
    \mathcal{B}(a),\; \\
    \mathcal{N}(\mathrm{series})
  \bigr\},
\end{align*}
where \(r,a,b>0\), \((x_1,y_1),(x_2,y_2),(h,k)\in\mathbb{R}^2\), \(a\neq 0\), and "series" ranges over the standard NACA four-digit codes \(\{2000,\dots,3000\}\).  Each shape is defined on \ref{fig:math_eq}.
\begin{figure*}[htbp!]
\centering
\begin{align*}
  \mathcal{C}(r,x_1,y_1)
  &= \bigl\{(x,y)\mid(x - x_1)^2 + (y - y_1)^2 = r^2\bigr\},\\
  \mathcal{E}(a,b,x_1,y_1)
  &= \bigl\{(x,y)\mid\tfrac{(x - x_1)^2}{a^2}+\tfrac{(y - y_1)^2}{b^2}=1\bigr\},\\
  \mathcal{L}(x_1,y_1,x_2,y_2)
  &= \Bigl\{(x,y)\mid y - y_1 = \tfrac{y_2 - y_1}{x_2 - x_1}(x - x_1)\Bigr\},\\
  \mathcal{P}(a,h,k)
  &= \bigl\{(x,y)\mid y = a\,(x - h)^2 + k\bigr\},\\
  \mathcal{B}(a)
  &= \bigl\{(x,y)\mid (x^2 + y^2)^2 = 2\,a^2\,(x^2 - y^2)\bigr\},\\
  \mathcal{N}(\mathrm{series})
  &= \bigl\{(x,y)\mid(x,y)\text{ satisfies the standard NACA upper/lower formulas}\bigr\}.
\end{align*}
\caption*{}
\phantomcaption
\label{fig:math_eq}
\end{figure*}

For each \( S\in\{\mathrm{circle, ellipse, line, parabola ,LB, NACA}\} \), we sample 5 independent instances by drawing all shape parameters (centres, radius, semi-axes, vertex offsets, scales, chord lengths, and series codes) uniformly from their prescribed domains.  We then choose \(n\in\{4\}\) uniformly at random and sample each curve at \(n\) parameter values to obtain \(\{(x_i,y_i)\}_{i=1}^n\), which serve as the target profiles.

\paragraph{Shape eligibility criteria.}
\label{app:shape_eligibility}
The six target classes were selected against four criteria. \textbf{(i) Analytic ground truth.} Each shape admits a closed-form equation, so a coupler path is correct or incorrect by construction. This is the property the symbolic interface relies on: \ac{sr} can recover an analytical surrogate and compare it to a known target, yielding verifiable rather than subjective feedback. \textbf{(ii) Curve-family diversity.} The set deliberately samples qualitatively different families, linear (line), conic (circle, ellipse, parabola), transcendental quartic (\ac{lb}), and an engineered piecewise-analytic profile (NACA), together covering open and closed, bounded and unbounded, single- and multi-lobe, and symmetric and asymmetric curves. \textbf{(iii) Established targets.} All six are conventional path-generation targets in the mechanism-synthesis literature \cite{Garc_a_Marina_2020,Tsai1999,Romero2019}, which keeps results comparable to prior work. \textbf{(iv) Emergent, kinematic difficulty.} We did not order the shapes by algebraic degree; the difficulty is an emergent property of kinematic feasibility. Table~\ref{tab:complex} orders mean Chamfer distance as NACA $<$ ellipse $<$ line $<$ circle $<$ \ac{lb} $<$ parabola, an ordering that algebraic degree does not predict: the parabola is a degree-two conic yet is hardest by two orders of magnitude, because it is open and unbounded and cannot be matched by the closed curve a coupler point traces (the corresponding critique reasoning is shown in App.~\ref{app:critique_parabola}), whereas the most engineered profile, the NACA airfoil, is smooth, bounded, and easiest. Shapes without a closed-form expression, such as free-hand or purely data-defined curves, were therefore excluded: they provide no analytic ground truth for \ac{sr} to recover and no rigorous correctness signal, which is exactly the property the method depends on.

\paragraph{Lemniscate of Bernoulli}
The lemniscate of Bernoulli is a figure-eight-shaped curve defined by the quartic equation $(x^2 + y^2)^2 = 2\,a^2\,(x^2 - y^2)$, where the parameter $a$ controls its overall size. It is symmetric about both axes and consists of two lobes meeting at the origin, exemplifying a simple rational algebraic curve.

\paragraph{NACA Four-Digit Airfoils}
Each NACA four-digit airfoil is specified by a code (e.g., "2412") indicating maximum camber, camber position, and thickness relative to chord length. Upper and lower surface coordinates are generated by piecewise analytic formulas that ensure smooth curvature and typical aerodynamic properties, making them standard in preliminary wing design.




\section{Appendix: \acf{da} Prompt}
\label{sec:dap}

\pageref{promtp:dap} serves as the template for the \ac{da} Prompt. It provides the structure and context that guides the \ac{da} in understanding how to approach and execute design of a planar mechanism.

\begin{nolinenumbers}
\begingroup
    \texttt{
    \textcolor{brown}{You are an AI specialized in designing planar mechanisms. Based on the description, generate the appropriate mechanism using Python with the pylinkage library} and \textcolor{violet}{explain each component step by step}.\\[1mm]
    \textcolor{red}{Commands (API Documentation): \{api\_doc\}}\\[1mm]
    \textcolor{blue}{Examples:}\\[1mm]
    \textcolor{blue}{\# Example 1}\\
    \textcolor{blue}{import pylinkage as pl}\\
    \textcolor{blue}{crank = pl.Crank(x=0, y=0, joint0=(0, 0), distance=1, angle=0.1, name="Crank")}
    }
    \texttt{
    \textcolor{blue}{slider = pl.Linear(x=2, y=0, joint0=crank, joint1=(0, 0), joint2=(1, 0), revolute\_radius=1.5, name="target")}\\[2mm]
    \textcolor{cyan}{\{memory\}} \\
    \textcolor{cyan}{The analytical equation describing the motion of the target joint in the above code is given by: \{equation\}
    The Chamfer distance of the target equation in the above code is: \{score\} \\[1mm]
    Our goal is to minimise the distance. Therefore, the greater the distance, the more it is not following the target motion and deviating from the intended path.} \\
    Planar Mechanism Description:\\
    \textcolor{red}{\{description\}}\\[2mm]
    The mechanism must pass as close as possible through all these points: \textcolor{green}{\{points\}}\\[2mm]
    Target analytical equation of the motion of the target joint: \textcolor{orange}{\{target\_equation\}}\\[2mm]
    \textcolor{purple}{Planar Mechanism Code:}\\[2mm]
    }
\label{promtp:dap}
\endgroup
\end{nolinenumbers}

\textbf{Legend for Colour Coding:}
\begin{itemize}
    \item \textcolor{brown}{\textbf{Brown}}: Agent Role.
    \item \textcolor{violet}{\textbf{Violet}}: Chain-of-Thoughts (CoT).
    \item \textcolor{red}{\textbf{Red}}: API Documentation and Mechanism Descriptions.
    \item \textcolor{blue}{\textbf{Blue}}: Examples and code snippets.
    \item \textcolor{cyan}{\textbf{Cyan}}: Memory: previously retained knowledge.
    \item \textcolor{green}{\textbf{Green}}: Key points or constraints (e.g., points through which the mechanism must pass).
    \item \textcolor{orange}{\textbf{Orange}}: Target analytical equation defining ideal behavior.
    \item \textcolor{purple}{\textbf{Purple}}: Final mechanism code output.
\end{itemize}

\section{Appendix: \acf{ca} Prompt}
\label{sec:cap}

\pageref{promtp:cap} acts as the template for the \ac{ca} Prompt. It outlines the format and expectations that help the \ac{ca} deliver clear, structured, and constructive critiques to support design improvement and decision-making.

\begin{nolinenumbers}
\begingroup
    \texttt{
    \textcolor{brown}{You are a reviewer for a mechanical designer AI agent.}\\[1mm]
    The following planar mechanism description: \textcolor{red}{\{description\}}\\[1mm]
    Simulator Output: \textcolor{orange}{\{simulator\_output\}}\\[1mm]
    \textcolor{cyan}{\{memory\}}\\[1mm]
    \textcolor{cyan}{The analytical equation describing the motion of the target joint in the above code is given by: \{equation\}
    The Chamfer distance of the target equation in the above code is: \{score\} \\[1mm]
    Our goal is to minimise the distance. Therefore, the greater the distance, the more it is not following the target motion and deviating from the intended path.} \\
    The following response was generated to fulfill the planar mechanism description:\\[1mm]
    Response: \textcolor{purple}{\{designer\_response\}}
    } \\
    \texttt{
    \textcolor{teal}{Your task is to evaluate the correctness, completeness, and complexity of the designed planar mechanism.  \\[1mm]
    Check for consistency with the problem constraints and point out any errors or improvements needed.  \\[1mm]
    Evaluate the complexity of the mechanism design in terms of:
    Structural Complexity: Assess whether the design is overly complex or can be simplified while maintaining functionality. \\[1mm]
    }
    \texttt{
    \textcolor{teal}{Structural Complexity: Assess whether the design is overly complex or can be simplified while maintaining functionality. \\[1mm]
    Design Elegance: Consider whether the design achieves the required functionality with minimal components or steps, adhering to principles of simplicity and elegance.\\ [1mm]
    Provide feedback in plain text. Point out areas where complexity could be reduced, and suggest improvements if necessary.}  \\[1mm]}
    }
\label{promtp:cap}
\endgroup
\end{nolinenumbers}

\textbf{Legend for Colour Coding:}
\begin{itemize}
    \item \textcolor{brown}{\textbf{Brown}}: Agent's Role.
    \item \textcolor{red}{\textbf{Red}}: Planar mechanism description.
    \item \textcolor{orange}{\textbf{Orange}}: Simulator output.
    \item \textcolor{cyan}{\textbf{Cyan}}: Memory: previously retained knowledge.
    \item \textcolor{purple}{\textbf{Purple}}: Designer's response.
    \item \textcolor{teal}{\textbf{Teal}}: Agent's task.
\end{itemize}

\section{Appendix: \acf{da} Revision Prompt}
\label{sec:darp}

\pageref{promtp:darp} serves as the template for the \ac{da} Revision Prompt. It provides the structure and context that guides the \ac{da} in understanding how to approach and execute design revisions effectively.

\begin{nolinenumbers}
\begingroup
    \texttt{
    You previously generated the following response for a planar mechanism description: \textcolor{purple}{\{designer\_response\}}\\[1mm]
    The reviewer provided the following feedback: \textcolor{magenta}{\{critique\_response\}}\\[1mm]
    Simulator Output: \textcolor{orange}{\{simulator\_output\}}\\[1mm]
    \textcolor{cyan}{\{memory\}}\\[1mm]
    \textcolor{cyan}{The analytical equation describing the motion of the target joint in the above code is given by: \{equation\}
    The Chamfer distance of the target equation in the above code is: \{score\} \\[1mm]
    Our goal is to minimise the distance. Therefore, the greater the distance, the more it is not following the target motion and deviating from the intended path.} \\
    \textcolor{teal}{Please revise your response to address the feedback and improve the planar mechanism.\\[1mm]
    The model should structure the response ensuring each step has only one line of code, ensuring clarity and logical progression, strictly adhering to the commands provided.}\\[1mm]
    Planar Mechanism Code:
    }
\label{promtp:darp}
\endgroup
\end{nolinenumbers}

\textbf{Appendix: Legend for Colour Coding:}
\begin{itemize}
    \item \textcolor{purple}{\textbf{Purple}}: Previously generated response (initial planar mechanism code).
    \item \textcolor{magenta}{\textbf{Magenta}}: Reviewer feedback.
    \item \textcolor{orange}{\textbf{Orange}}: Simulator output.
    \item \textcolor{cyan}{\textbf{Cyan}}: Memory: previously retained knowledge.
    \item \textcolor{teal}{\textbf{Teal}}: Agent's task.
\end{itemize}

\section{Appendix: Shape Complexity}
\label{sec:shape_complex}

In this section, we present metrics for quantifying shape complexity and compare them to our ground truth and agent-generated forms (Table~\ref{tab:complex}). This concise analysis illustrates how fidelity deteriorates (or remains stable) as geometric intricacy increases. By comparing complexity scores directly across shapes, we can identify the limits of our agent's performance.

\begin{table*}[!h]
	\centering
	\begin{tabular}{@{}lc@{}}
		\toprule
		Shape & Mean Chamfer distance \\
		\midrule
        \ac{lb} & $8.5654 \pm 0.1632$ \\
        Circle & $7.4564 \pm 0.5147$ \\
        Ellipse & $5.4398 \pm 0.1789$ \\
        Line & $7.1046 \pm 0.5392$ \\
        NACA & $1.1684 \pm 0.0659$ \\
        Parabola & $853.2922 \pm 6.7232$ \\
		\bottomrule
	\end{tabular}
    \caption{The mean Chamfer distances ($\pm$ standard error) are shown for each shape category, calculated from the motion of the contours generated by the planar mechanism and the ground-truth contours.}
    \label{tab:complex}
\end{table*}

\section{Appendix: SR-Enabled vs.\ SR-Absent Critique Comparison}
\label{app:critique_comparison}

\begin{figure*}[t]
    \centering
    \includegraphics[width=0.95\textwidth]{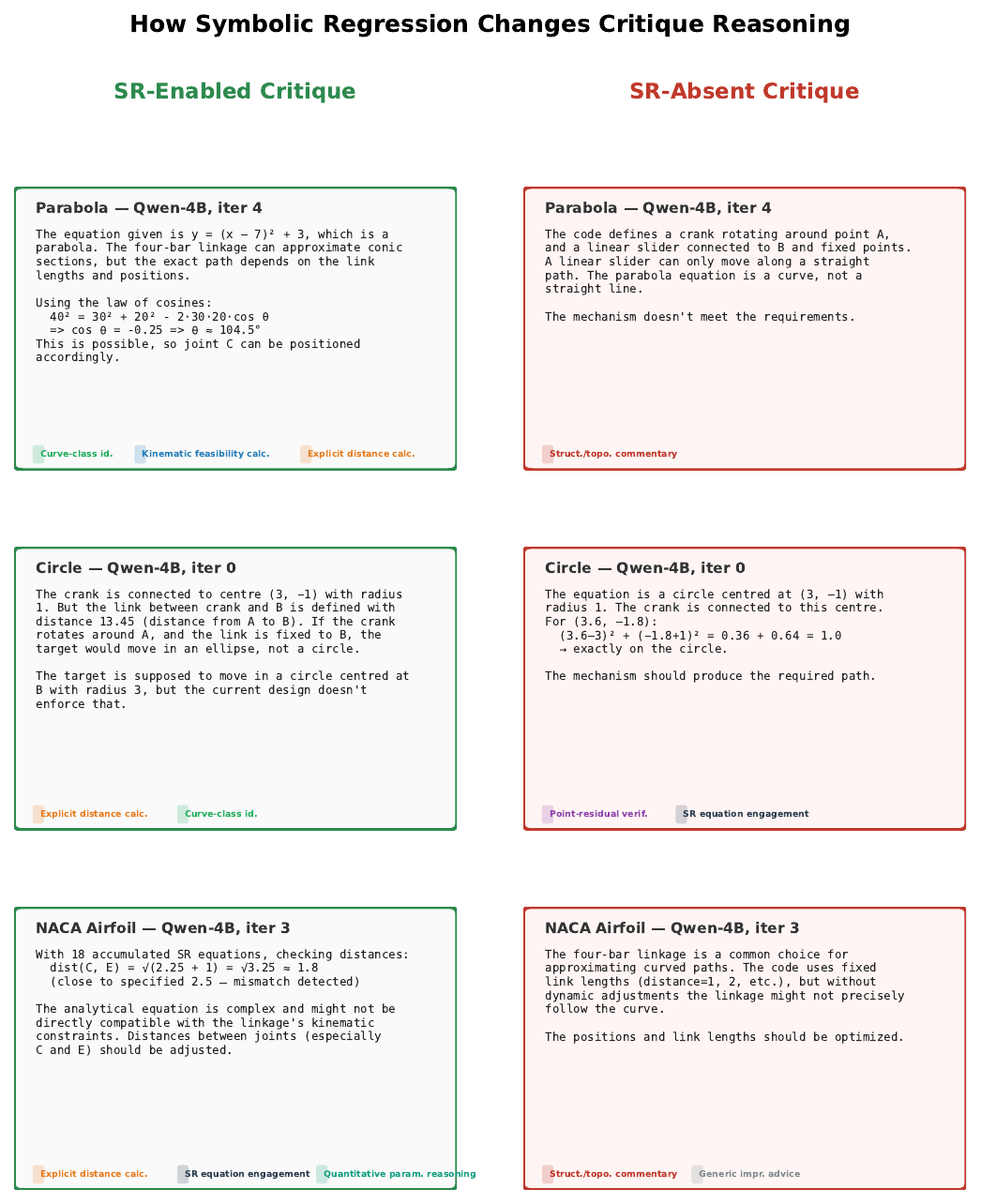}
    \caption{Paired critique comparison showing how symbolic regression changes the nature of reasoning. \textbf{Left}: SR-enabled critiques identify curve classes, apply kinematic calculations (law of cosines), and compute explicit distances. \textbf{Right}: matched SR-absent critiques default to structural heuristics and generic improvement advice. Coloured tags indicate the reasoning types present.}
    \label{fig:qualitative_comparison_app}
\end{figure*}

The following paired excerpts illustrate how the symbolic regression bridge changes the character of the critique agent's reasoning. In each pair, the same model and target shape are used; only the availability of \ac{sr}-fitted equations differs. Excerpts are lightly abridged for space; full transcripts are available in the supplementary material.

\subsection{Parabola - Qwen (4B), iteration 4}
\label{app:critique_parabola}

\noindent\textbf{SR-enabled critique} (\ac{sr} equations available; 15 accumulated expressions):\\[2pt]
\begin{quote}
\small\ttfamily
The equation given is $y = (x - 7)^2 + 3$, which is a parabola. The points provided (like $(32.56,\,656.14)$) lie on this curve. The mechanism should be designed to approximate this path. [\ldots] Now, I need to verify if this setup can produce a parabolic path. The movement of joint B (the crank) is circular around A. As B rotates, C moves in a circle around D. The target E is connected to both B and C, so its position is determined by the relative positions of B and C.

Does this configuration result in E tracing a parabola? The parabola is a quadratic curve, which is a second-order curve. A four-bar linkage can approximate conic sections, but the exact path depends on the link lengths. [\ldots] The distance between B and D is $\sqrt{40^2 + 20^2} = \sqrt{2000} \approx 44.72$. Using the law of cosines: $40^2 = 30^2 + 20^2 - 2 \cdot 30 \cdot 20 \cos\theta \Rightarrow \cos\theta = -0.25 \Rightarrow \theta \approx 104.48^{\circ}$. This is possible, so C can be positioned accordingly.
\end{quote}

\noindent\textbf{SR-absent critique} (no \ac{sr} equations):\\[2pt]
\begin{quote}
\small\ttfamily
The code defines a crank (B) rotating around point A (7,3), and a linear slider (C) connected to B and fixed points A and D (14,3). [\ldots] A linear joint is typically a slider that moves along a straight line. If A and D are at $(7,3)$ and $(14,3)$, the line AD is horizontal. So the slider C would move along this horizontal line. However, the parabola equation is a curve, not a straight line. This seems contradictory. [\ldots] The design is incorrect. The linear joint can't produce a parabola. [\ldots] The mechanism doesn't meet the requirements.
\end{quote}

\noindent\textbf{Key difference.}
The SR-enabled critique identifies the curve class (parabola), computes whether the four-bar linkage can kinematically realise it via the law of cosines, and evaluates link-length compatibility, all geometric reasoning mediated by the symbolic representation. The SR-absent critique detects a topological error (linear slider vs.\ curved path) but does not engage analytically with the curve geometry or propose quantitative corrections.

\subsection{NACA Airfoil - Qwen (4B), iteration 3}
\label{app:critique_naca}

\noindent\textbf{SR-enabled critique} (18 accumulated \ac{sr} equations):\\[2pt]
\begin{quote}
\small\ttfamily
The static points A and D are at $(0,0)$ and $(3,0)$. [\ldots] Joint C is at $(3,2)$, with distances of 3 from B and 1.5 from D. If B is at $(0,1)$ and C is at $(3,2)$, the distance between B and C would be $\sqrt{(3{-}0)^2 + (2{-}1)^2} = \sqrt{10} \approx 3.16$, which is close to the specified 3 units. [\ldots] The analytical equation is complex and might not be directly compatible with the linkage's kinematic constraints. [\ldots] The distances between joints (especially between C and E) could be adjusted to better approximate the desired path.
\end{quote}

\noindent\textbf{SR-absent critique} (no \ac{sr} equations):\\[2pt]
\begin{quote}
\small\ttfamily
The four-bar linkage is a common choice for approximating curved paths, especially when combined with specific link lengths and angles. [\ldots] The code uses fixed link lengths (distance=1, 2, etc.), but without dynamic adjustments, the linkage might not be able to precisely follow the curve. [\ldots] The positions of the fixed points and link lengths should be optimized to align with these points.
\end{quote}

\noindent\textbf{Key difference.}
The SR-enabled critique performs explicit distance verification ($\sqrt{10} \approx 3.16$ vs.\ specified 3), flags incompatibility between the \ac{sr}-fitted analytical equation and the linkage topology, and localises the source of error to a specific joint pair (C--E). The SR-absent critique offers generic optimisation advice without quantitative grounding.

\subsection{Circle - Qwen (4B), iteration 0}
\label{app:critique_circle}

\noindent\textbf{SR-enabled critique}:\\[2pt]
\begin{quote}
\small\ttfamily
The crank is connected to this centre, and the target is at the end of the crank. As the crank rotates, the target should move in a circle around the centre. The crank's length is 1, which matches the radius. [\ldots] But the link between the crank and B is defined with a distance of 13.45, which is the distance from A to B. If the crank is rotating around A, and the link is fixed between the crank's end and B, then the target would move in an ellipse, not a circle.
\end{quote}

\noindent\textbf{SR-absent critique}:\\[2pt]
\begin{quote}
\small\ttfamily
The equation given is a circle centred at $(3, {-}1)$ with radius 1. The crank is connected to this centre, and the target is at the end of the crank. [\ldots] For $(3.6,{-}1.8)$: $(3.6{-}3)^2 + ({-}1.8{+}1)^2 = 0.36 + 0.64 = 1.0 \to$ exactly on the circle.
\end{quote}

\noindent\textbf{Key difference.}
In both conditions, the critique agent engages with the circle equation. However, the SR-absent run happened to receive a near-optimal design from the first iteration (Chamfer $= 0.003$), so the critique merely verifies correctness. The SR-enabled critique detects a subtler design flaw, a distance mismatch that would produce an ellipse rather than a circle, demonstrating that symbolic feedback supports \emph{error diagnosis} even on simple shapes.

\subsection{Quantitative reasoning taxonomy}
\label{app:reasoning_taxonomy}

To complement the paired excerpts, we programmatically scored all $30{,}977$ critique entries with the eight-category, splitting by model and \ac{sr} condition.

\paragraph{The robust cross-model signal.} Engagement with discovered symbolic structure rises with \ac{sr} for \emph{every} model (\ac{gemma} $0\%\!\to\!2\%$, \ac{llama} $1\%\!\to\!3\%$, \ac{qwen} $16\%\!\to\!26\%$): the interface measurably changes \emph{what} critiques reason over, independently of architecture.

\paragraph{Grounding, not frequency: a compositional caveat.} A na\"ive aggregate reading suggests \ac{sr}-absent critiques are \emph{more} analytical (e.g.\ explicit distance calculation in $19.4\%$ of \ac{sr}-off vs.\ $19.2\%$ of \ac{sr}-on entries, and a higher overall grounded-analytical share). This aggregate pattern is a \emph{compositional} (Simpson) effect: the \ac{sr}-off pool is dominated by \ac{llama}~70B, which emits analytical-sounding language at high frequency regardless of whether it connects to the target geometry (its grounded-analytical share is only $25$--$32\%$). Conditioning on model removes the paradox: the reasoning-trained compact model \ac{qwen}~4B is precisely where \ac{sr} raises grounded analytical reasoning ($96\%\!\to\!98\%$), while the larger dense models do not increase it. The effect is therefore one of \emph{grounding} anchored in discovered equations, an architecture-dependent shift (consistent with \S\ref{sec:discussion}, ``Why architecture matters''), which aggregate metrics alone cannot capture.

\section{Appendix: Emergent curriculum}
\label{app:curriculum}

Section~\ref{sec:discussion} claims the critique loop induces an \emph{emergent curriculum}: structural commentary dominates early and gives way to parametric refinement. We test this directly. Over $180$ \ac{sr}- and feedback-enabled experiments ($443{,}604$ critique sentences), the share of sentences flagged \emph{structural} (topology, joint changes, redesign) falls from $5.69\%$ at iteration~$0$ to $3.39\%$ at iteration~$19$, while \emph{parametric} sentences (link-length, position, angle adjustments) stay essentially flat ($\approx\!4.9\%$); the structural-to-parametric ratio declines from $1.08$ to $0.66$. A $\chi^2$ test of structural sentence frequency, early vs.\ late iterations, is highly significant ($\chi^2\!=\!278.67$, $p\!\approx\!1.5\!\times\!10^{-62}$), whereas the parametric change is only marginal ($\chi^2\!=\!4.23$, $p\!=\!0.04$): difficulty shifts from \emph{what topology} to \emph{which parameters} without any explicit curriculum design. Representative early-structural / late-parametric pairs from the same run:

\begin{quote}\small
\textbf{Gemma 12B, \ac{lb}.} \emph{Iter 4 (structural):} ``the simulation output indicates a critical issue: \texttt{UnbuildableError} [\ldots] the underlying mechanism design is fundamentally flawed.'' $\rightarrow$ \emph{Iter 16 (parametric):} ``Re-evaluate joint positions \& link lengths: the $x,y$ coordinates of joints C and D, and the \texttt{distance0}, \texttt{distance1} values, need to be carefully calculated.''

\smallskip
\textbf{Gemma 12B, circle.} \emph{Iter 2 (structural):} ``Prioritise a four-bar linkage: the most significant improvement would be to restructure the mechanism.'' $\rightarrow$ \emph{Iter 17 (parametric):} ``Optimise link lengths: incorporate an optimisation routine to adjust the link lengths.''
\end{quote}

\section{Appendix: Qualitative Results}
\label{sec:qua}

\begin{figure}[!htb]
\renewcommand\thefigure{5}
\includegraphics[width=0.8\columnwidth]{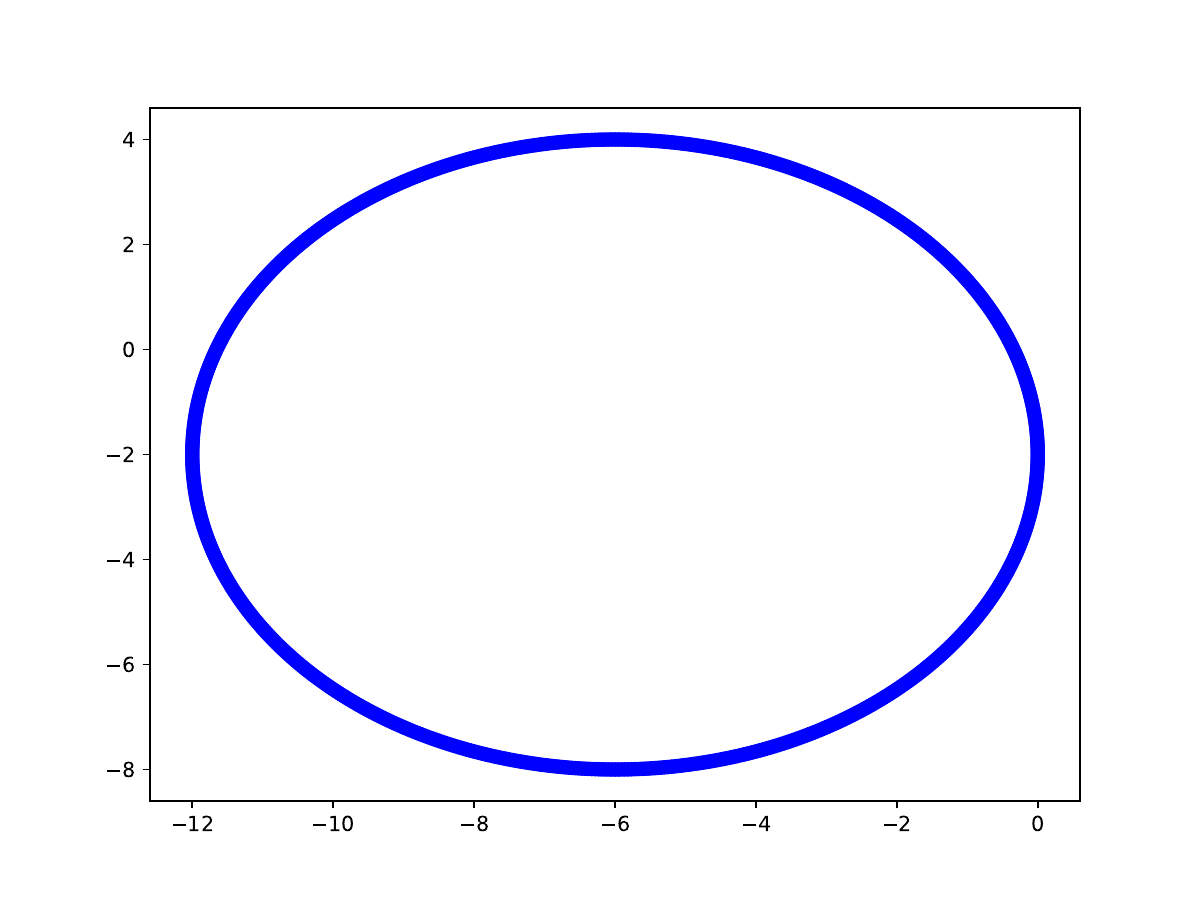}
  \caption{Ground truth Ellipse: the original target Ellipse used for evaluation.}
  \label{fig:target1}
\end{figure}

\begin{figure}[!htb]
\renewcommand\thefigure{6}
  \includegraphics[width=0.8\columnwidth]{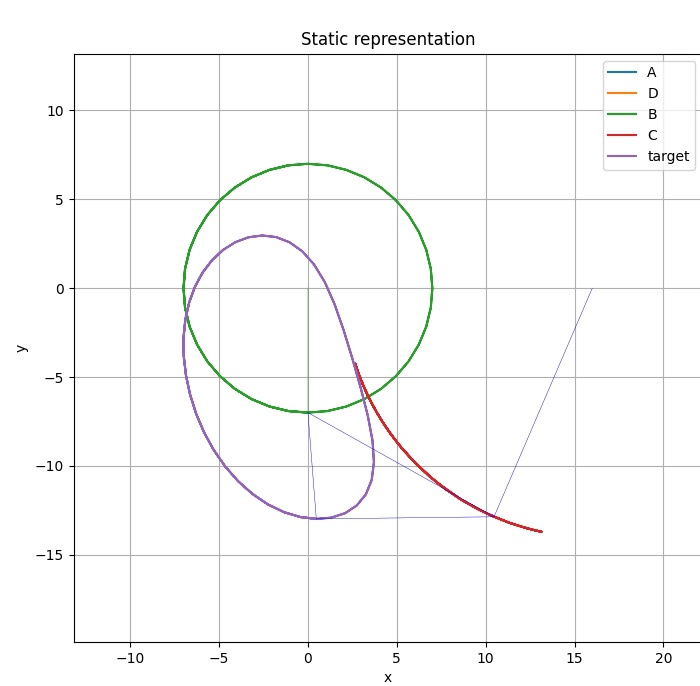}
  \caption{Generated Ellipse: the Ellipse synthesised by our agent.}
  \label{fig:experiments1}
\end{figure}

\begin{figure}[!htb]
\renewcommand\thefigure{7}
\includegraphics[width=0.9\columnwidth]{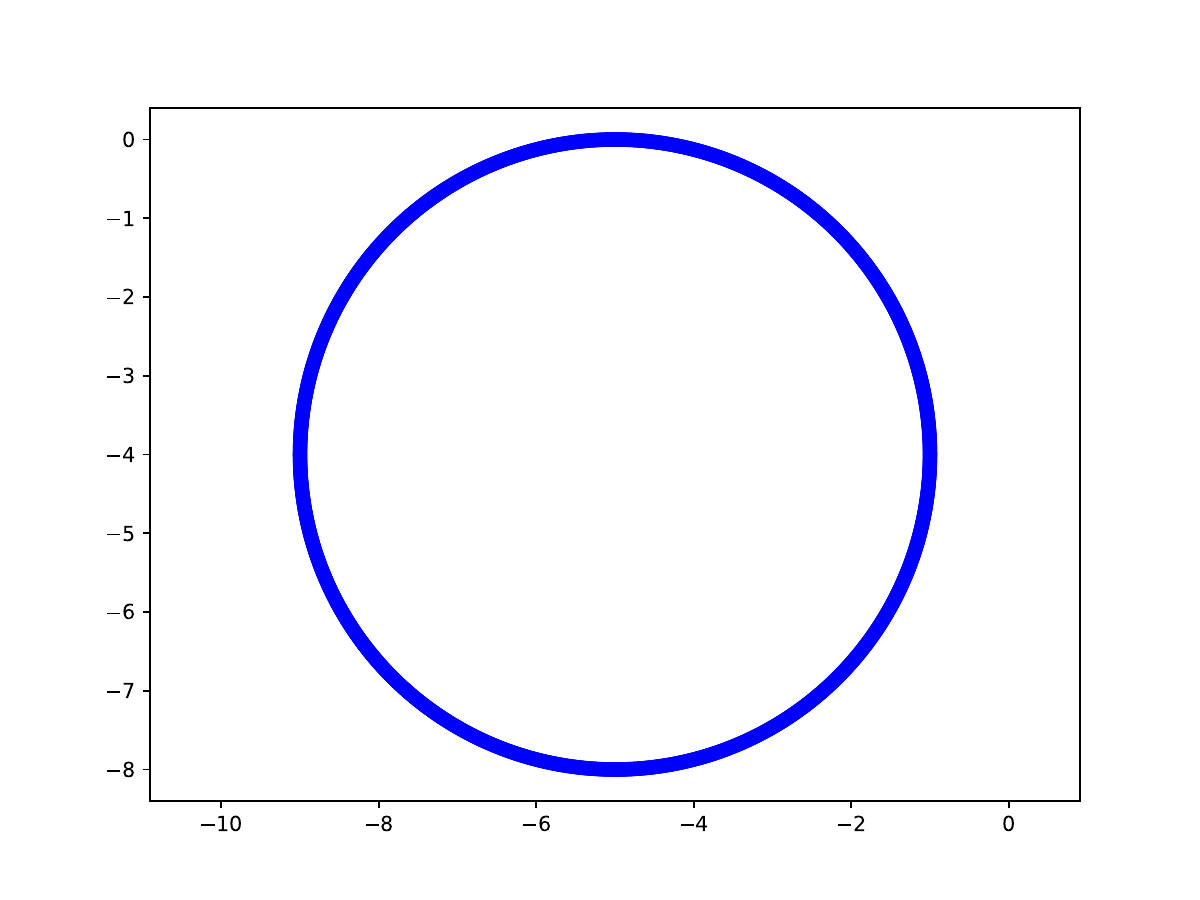}
  \caption{Ground truth Circle: the original target Circle used for evaluation.}
  \label{fig:target2}
\end{figure}

\begin{figure}[!htb]
\renewcommand\thefigure{8}
  \includegraphics[width=0.9\columnwidth]{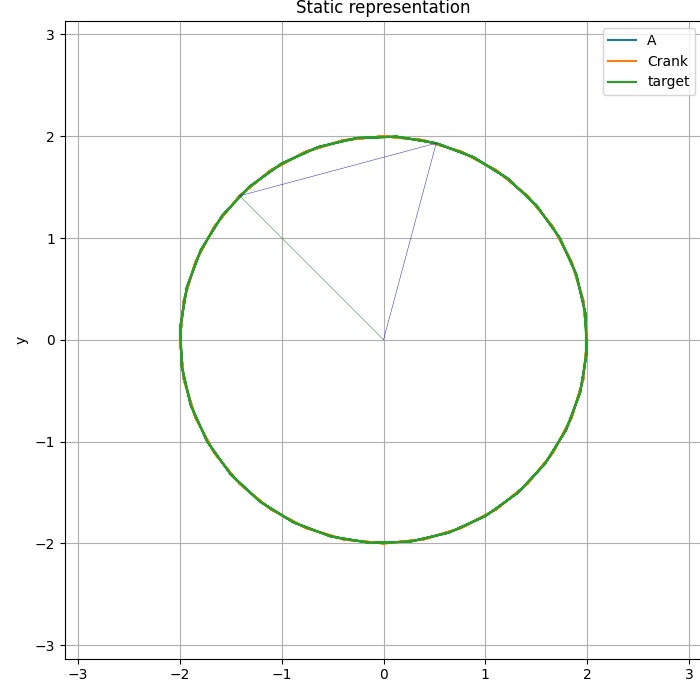}
  \caption{Generated Circle: the Circle synthesised by our agent.}
  \label{fig:experiments2}
\end{figure}

\begin{figure}[!htb]
\renewcommand\thefigure{9}
  \includegraphics[width=0.6\columnwidth]{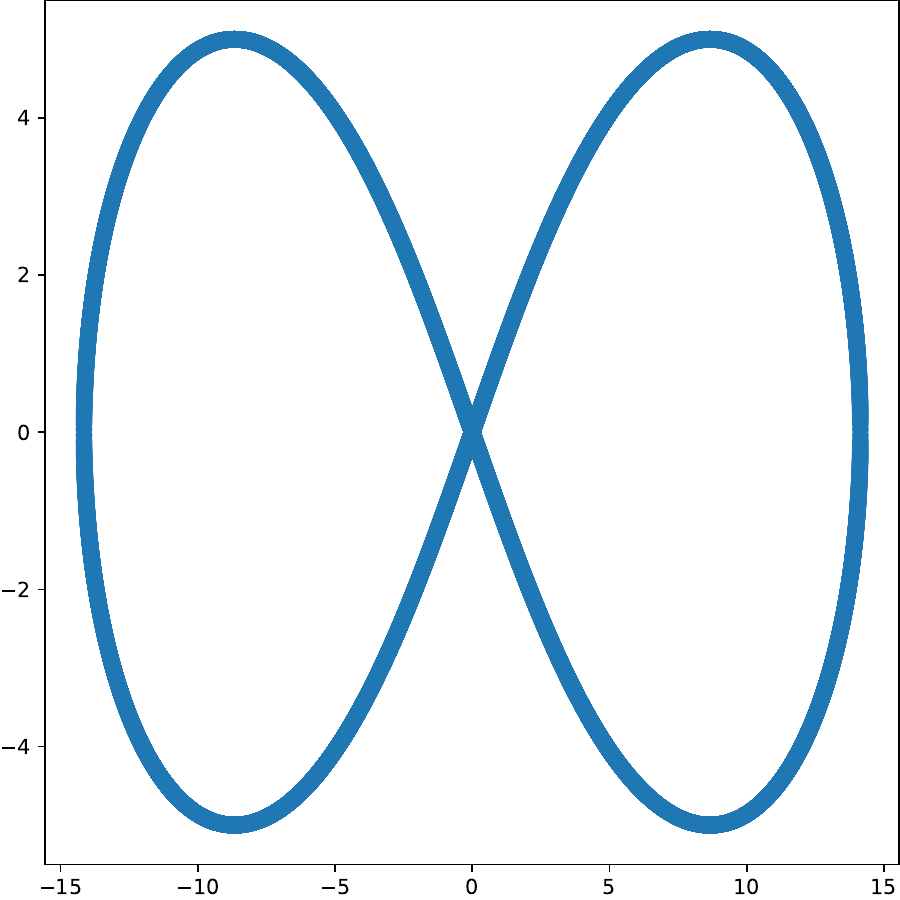}
  \caption{Ground truth \ac{lb}: the original target \ac{lb} used for evaluation.}
  \label{fig:target3}
\end{figure}

\begin{figure}[!htb]
\renewcommand\thefigure{10}
  \includegraphics[width=0.9\columnwidth]{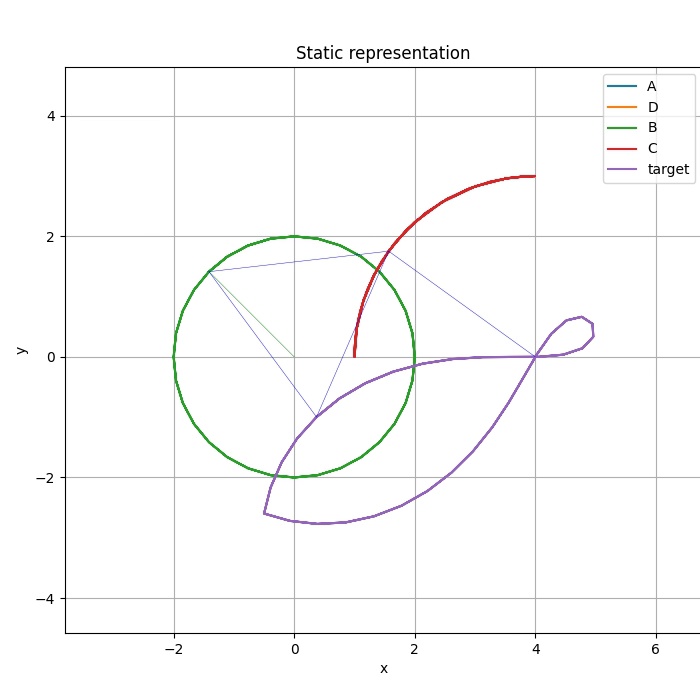}
  \caption{Generated \ac{lb}: the \ac{lb} synthesised by our agent.}
  \label{fig:experiments3}
\end{figure}

\begin{figure}[!htb]
\renewcommand\thefigure{11}
  \includegraphics[width=\columnwidth]{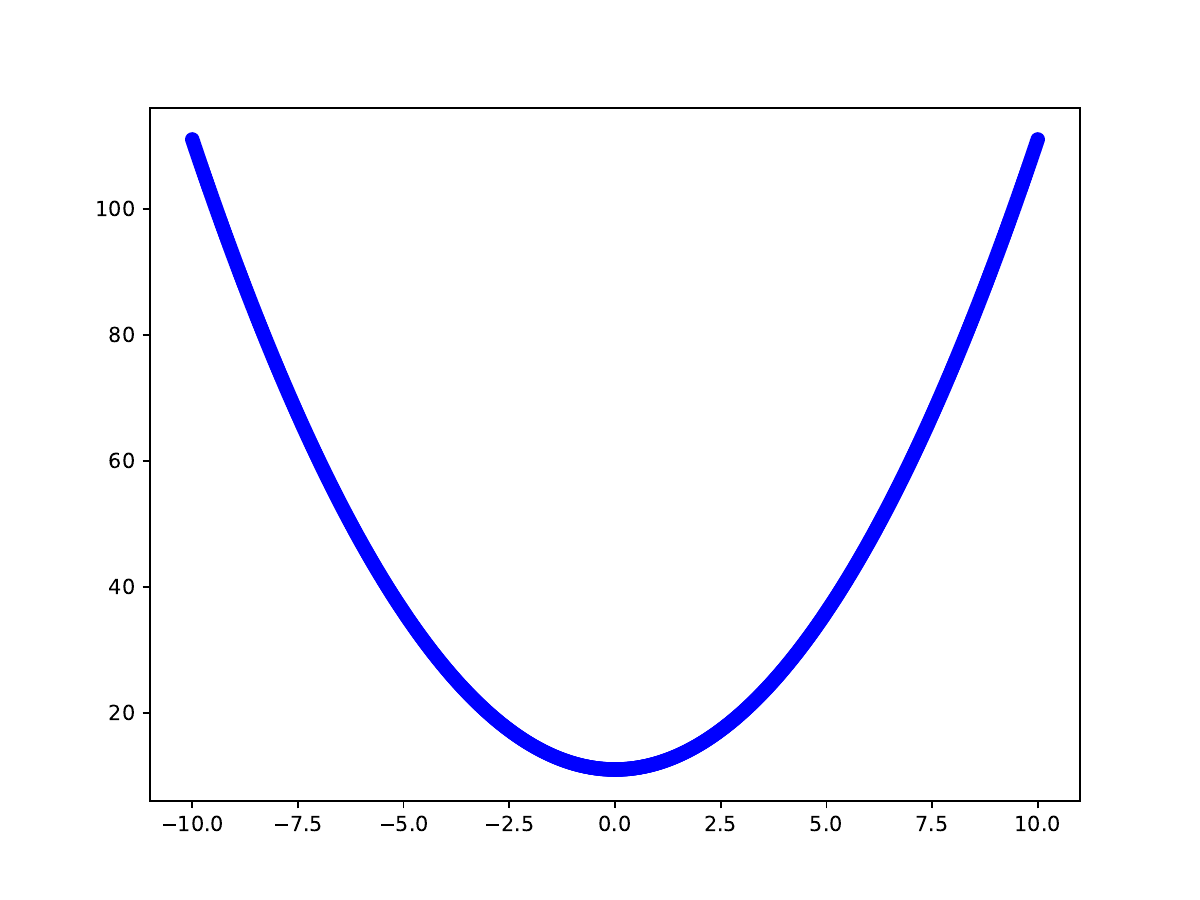}
  \caption{Ground truth Parabola: the original target Parabola used for evaluation.}
  \label{fig:target4}
\end{figure}

\begin{figure}[!htb]
\renewcommand\thefigure{12}
  \includegraphics[width=\columnwidth]{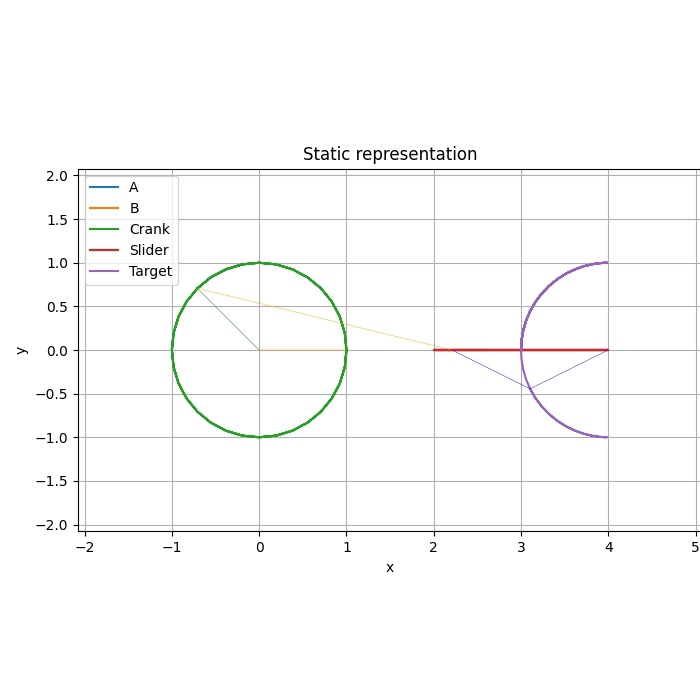}
  \caption{Generated Parabola: the Parabola synthesised by our agent.}
  \label{fig:experiments4}
\end{figure}

\begin{figure}[!htb]
\renewcommand\thefigure{13}
  \includegraphics[width=\columnwidth]{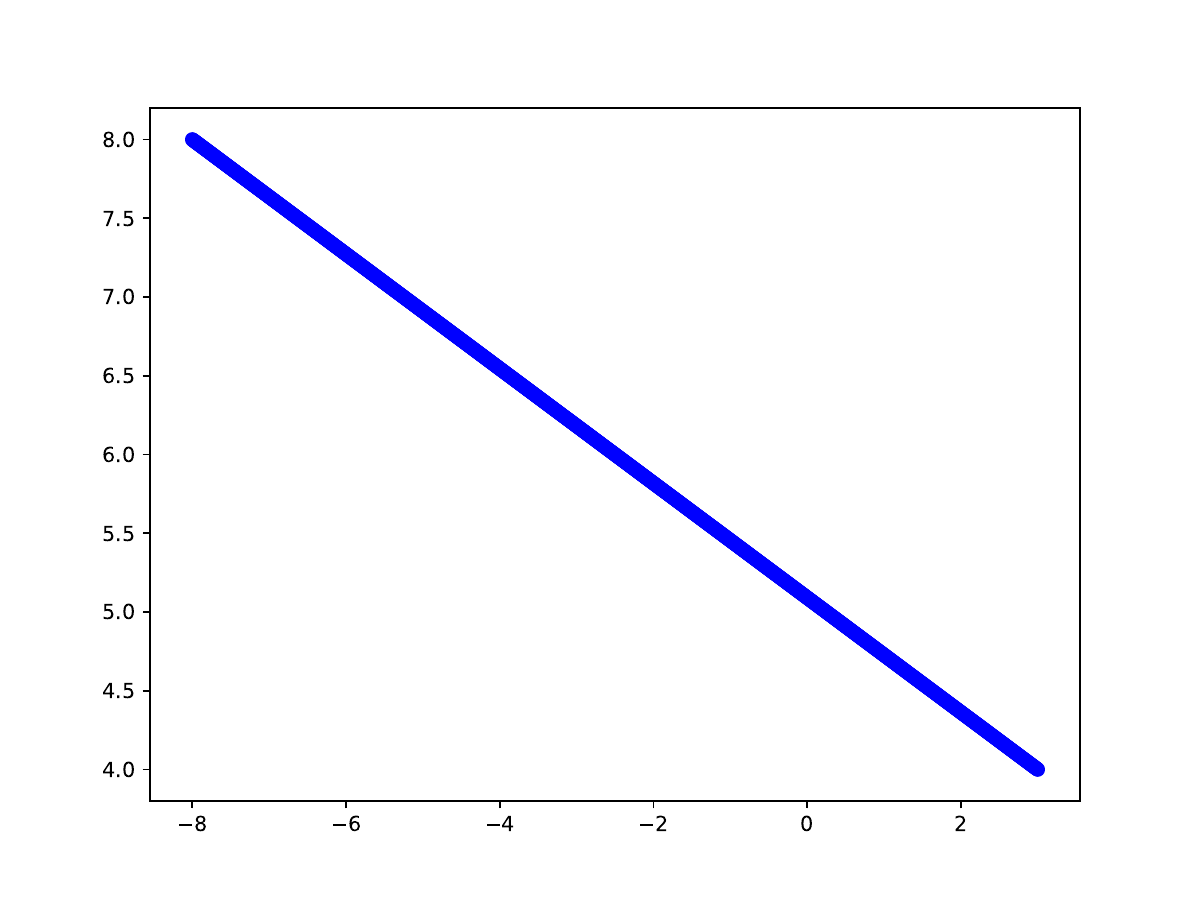}
  \caption{Ground truth Line: the original target Line used for evaluation.}
  \label{fig:target5}
\end{figure}

\begin{figure}[!htb]
\renewcommand\thefigure{14}
  \includegraphics[width=\columnwidth]{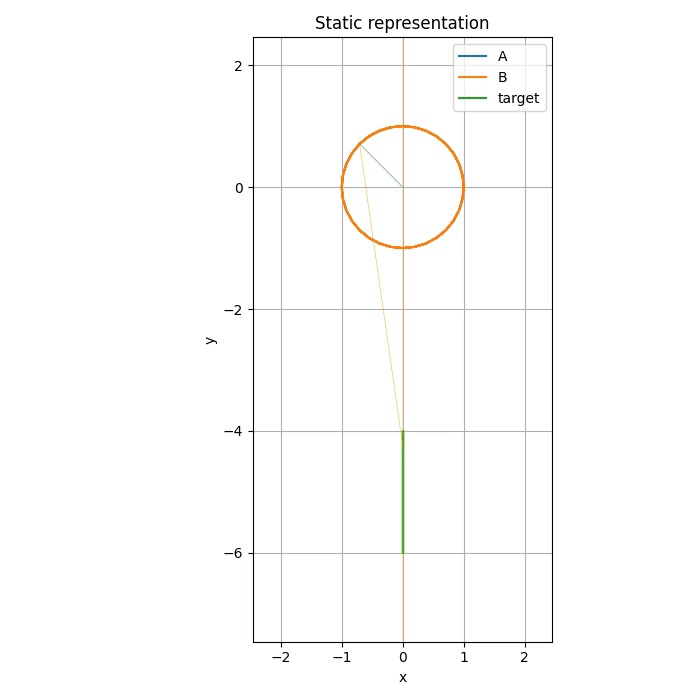}
  \caption{Generated Line: the Line synthesised by our agent.}
  \label{fig:experiments5}
\end{figure}

\begin{figure}[!htbp]
\renewcommand\thefigure{15}
  \includegraphics[width=\columnwidth]{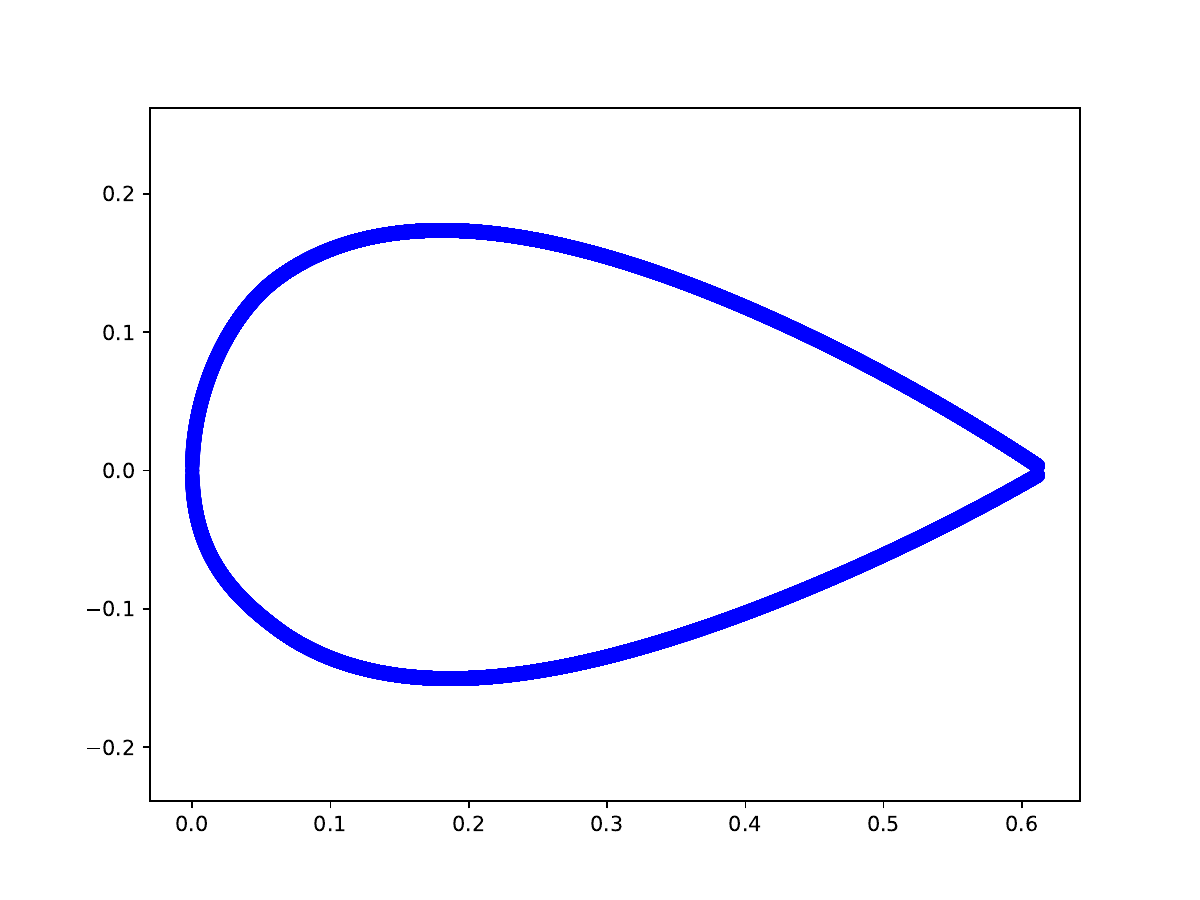}
  \caption{Ground truth NACA airfoil: the original target NACA airfoil used for evaluation.}
  \label{fig:target6}
\end{figure}

\begin{figure}[!htbp]
\renewcommand\thefigure{16}
  \includegraphics[width=0.8\columnwidth]{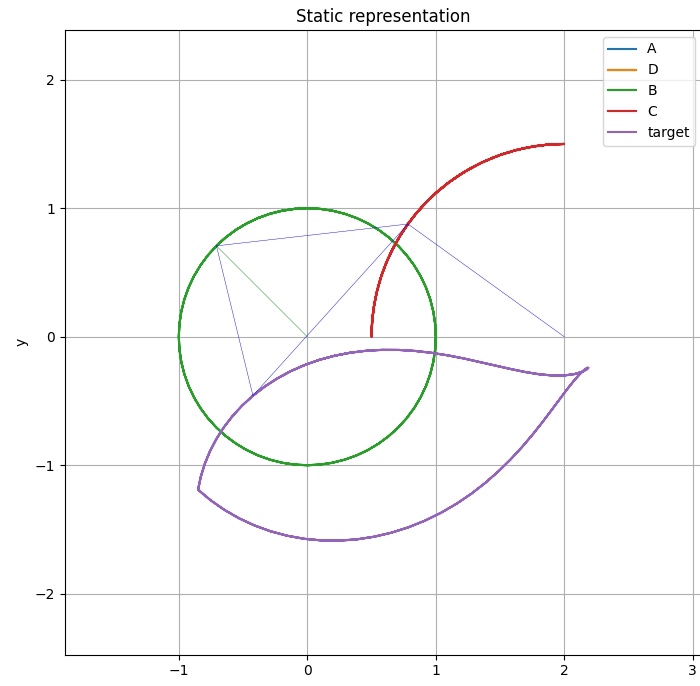}
  \caption{Generated NACA airfoil: the NACA airfoil synthesised by our agent.}
  \label{fig:experiments6}
\end{figure}

Figures \ref{fig:target1},\ref{fig:target2},\ref{fig:target3},\ref{fig:target4},\ref{fig:target5}, and \ref{fig:target6}, shows the ground-truth targets, while Figures \ref{fig:experiments1}, \ref{fig:experiments2},\ref{fig:experiments3}, \ref{fig:experiments4}, \ref{fig:experiments5}, and \ref{fig:experiments6} shows the corresponding outputs produced by our agent.  Close inspection reveals that many of the generated results closely align with their targets, accurately capturing both the global structure and fine-scale details.  This strong visual correspondence highlights the effectiveness of our method in approximating the desired outputs across a range of diverse test cases.

\section{Appendix: Hardware Configuration}

We used a workstation equipped with five NVIDIA RTX 3090 graphics cards, each with 24 GB of VRAM and 72 GB of system RAM.

\section{Appendix: Extended Quantitative Result}
\label{sec:all_quant}
In this section, we present the outcomes of our complete set of experiments, summarising performance across all six shape categories.

\subsection{Full Enum+GA Baseline Comparison}
Table~\ref{tab:model_vs_ga_full} compares the three \ac{llm} agents (with Feedback enabled) against all six Enum+GA baseline configurations spanning exhaustive topology enumeration over two linkage types (4-bar, 6-bar) and three GA parameter-tuning budgets (Pop\,3/Gen\,20, Pop\,6/Gen\,20, Pop\,60/Gen\,300). Values are per-shape normalised Chamfer ratios (\ac{llm} median / GA median, averaged over shapes); values below~1 indicate the \ac{llm} outperforms the GA.

\paragraph{Why the Enum+GA baseline.}
\label{app:baseline_eligibility}
We chose exhaustive topology enumeration with genetic-algorithm parameter tuning for four reasons. \textbf{(i) Established standard.} Enumeration of kinematic topologies followed by population-based dimensional optimisation is the canonical approach to path synthesis in computational kinematics \cite{Garc_a_Marina_2020,Tsai1999,Romero2019,vasiliu2001dimensional}, so it is the comparator a domain expert would reach for. \textbf{(ii) Strong, complete comparator.} Enumerating both 4-bar and 6-bar topologies covers the same hypothesis space the agent explores, so the comparison is against a thorough search rather than a weak heuristic. \textbf{(iii) Budget control.} The GA budget is tunable, which lets us match the agent's evaluation budget exactly (Pop\,3 / Gen\,20, 60 evaluations) and also report a $150\times$ upper bound (Pop\,60 / Gen\,300, 18{,}000 evaluations); this isolates reasoning efficiency from raw compute. \textbf{(iv) Verified near-optimal.} Convergence is confirmed for every topology and shape (App.~\ref{app:ga_convergence}), so the baseline is a fair reference rather than an under-tuned one. We did not adopt neural one-shot generators or gradient-based methods as the primary baseline: they require large curated training corpora or are sensitive to initialisation and penalty weighting \cite{lee2024deep,singh2024new,nobari2022links}, and neither yields a training-free, budget-matched comparison. Since our method uses no precomputed database, Enum+GA is the appropriate matched comparator.

\begin{table*}[ht]
\centering
\setlength{\tabcolsep}{8pt}
\renewcommand{\arraystretch}{1.3}
\begin{tabular}{@{} l c l r r r @{}}
\toprule
\textbf{Baseline} & \textbf{SFB} & \textbf{Model} & \textbf{Value} & $\boldsymbol{\Delta}$\,\textbf{(vs Base)} & \textbf{Imp.\,\%} \\
\midrule
\grayrow  &  & \gemma & \textbf{0.818} & \textcolor{teal}{$-$0.182 $\uparrow$} & \textcolor{teal}{$+$18.2\% $\uparrow$} \\
\grayrow  &  & \llama & \textbf{0.461} & \textcolor{teal}{$-$0.539 $\uparrow$} & \textcolor{teal}{$+$53.9\% $\uparrow$} \\
\grayrow  & \multirow{-3}{*}{$\times$} & \qwen & \textbf{0.582} & \textcolor{teal}{$-$0.418 $\uparrow$} & \textcolor{teal}{$+$41.8\% $\uparrow$} \\
\cmidrule{2-6}
\grayrow  &  & \gemma & \textbf{0.804} & \textcolor{teal}{$-$0.196 $\uparrow$} & \textcolor{teal}{$+$19.6\% $\uparrow$} \\
\grayrow  &  & \llama & \textbf{0.467} & \textcolor{teal}{$-$0.533 $\uparrow$} & \textcolor{teal}{$+$53.3\% $\uparrow$} \\
\grayrow \multirow{-6}{*}{\makecell[l]{\textbf{4-bar}\\Pop:\,3 / Gen:\,20}} & \multirow{-3}{*}{$\checkmark$} & \qwen & \textbf{0.475} & \textcolor{teal}{$-$0.525 $\uparrow$} & \textcolor{teal}{$+$52.5\% $\uparrow$} \\
\midrule
 &  & \gemma & \textbf{1.219} & \textcolor{red}{$+$0.219 $\downarrow$} & \textcolor{red}{$-$21.9\% $\downarrow$} \\
 &  & \llama & \textbf{0.634} & \textcolor{teal}{$-$0.366 $\uparrow$} & \textcolor{teal}{$+$36.6\% $\uparrow$} \\
 & \multirow{-3}{*}{$\times$} & \qwen & \textbf{0.845} & \textcolor{teal}{$-$0.155 $\uparrow$} & \textcolor{teal}{$+$15.5\% $\uparrow$} \\
\cmidrule{2-6}
 &  & \gemma & \textbf{1.185} & \textcolor{red}{$+$0.185 $\downarrow$} & \textcolor{red}{$-$18.5\% $\downarrow$} \\
 &  & \llama & \textbf{0.659} & \textcolor{teal}{$-$0.341 $\uparrow$} & \textcolor{teal}{$+$34.1\% $\uparrow$} \\
\multirow{-6}{*}{\makecell[l]{\textbf{4-bar}\\Pop:\,6 / Gen:\,20}} & \multirow{-3}{*}{$\checkmark$} & \qwen & \textbf{0.688} & \textcolor{teal}{$-$0.312 $\uparrow$} & \textcolor{teal}{$+$31.2\% $\uparrow$} \\
\midrule
\grayrow  &  & \gemma & \textbf{3.043} & \textcolor{red}{$+$2.043 $\downarrow$} & \textcolor{red}{$-$204.3\% $\downarrow$} \\
\grayrow  &  & \llama & \textbf{1.247} & \textcolor{red}{$+$0.247 $\downarrow$} & \textcolor{red}{$-$24.7\% $\downarrow$} \\
\grayrow  & \multirow{-3}{*}{$\times$} & \qwen & \textbf{2.097} & \textcolor{red}{$+$1.097 $\downarrow$} & \textcolor{red}{$-$109.7\% $\downarrow$} \\
\cmidrule{2-6}
\grayrow  &  & \gemma & \textbf{3.072} & \textcolor{red}{$+$2.072 $\downarrow$} & \textcolor{red}{$-$207.2\% $\downarrow$} \\
\grayrow  &  & \llama & \textbf{1.351} & \textcolor{red}{$+$0.351 $\downarrow$} & \textcolor{red}{$-$35.1\% $\downarrow$} \\
\grayrow \multirow{-6}{*}{\makecell[l]{\textbf{4-bar}\\Pop:\,60 / Gen:\,300}} & \multirow{-3}{*}{$\checkmark$} & \qwen & \textbf{1.320} & \textcolor{red}{$+$0.320 $\downarrow$} & \textcolor{red}{$-$32.0\% $\downarrow$} \\
\midrule
 &  & \gemma & \textbf{0.980} & \textcolor{teal}{$-$0.020 $\uparrow$} & \textcolor{teal}{$+$2.0\% $\uparrow$} \\
 &  & \llama & \textbf{0.558} & \textcolor{teal}{$-$0.442 $\uparrow$} & \textcolor{teal}{$+$44.2\% $\uparrow$} \\
 & \multirow{-3}{*}{$\times$} & \qwen & \textbf{0.710} & \textcolor{teal}{$-$0.290 $\uparrow$} & \textcolor{teal}{$+$29.0\% $\uparrow$} \\
\cmidrule{2-6}
 &  & \gemma & \textbf{0.955} & \textcolor{teal}{$-$0.045 $\uparrow$} & \textcolor{teal}{$+$4.5\% $\uparrow$} \\
 &  & \llama & \textbf{0.565} & \textcolor{teal}{$-$0.435 $\uparrow$} & \textcolor{teal}{$+$43.5\% $\uparrow$} \\
\multirow{-6}{*}{\makecell[l]{\textbf{6-bar}\\Pop:\,3 / Gen:\,20}} & \multirow{-3}{*}{$\checkmark$} & \qwen & \textbf{0.601} & \textcolor{teal}{$-$0.399 $\uparrow$} & \textcolor{teal}{$+$39.9\% $\uparrow$} \\
\midrule
\grayrow  &  & \gemma & \textbf{1.154} & \textcolor{red}{$+$0.154 $\downarrow$} & \textcolor{red}{$-$15.4\% $\downarrow$} \\
\grayrow  &  & \llama & \textbf{0.634} & \textcolor{teal}{$-$0.366 $\uparrow$} & \textcolor{teal}{$+$36.6\% $\uparrow$} \\
\grayrow  & \multirow{-3}{*}{$\times$} & \qwen & \textbf{0.829} & \textcolor{teal}{$-$0.171 $\uparrow$} & \textcolor{teal}{$+$17.1\% $\uparrow$} \\
\cmidrule{2-6}
\grayrow  &  & \gemma & \textbf{1.112} & \textcolor{red}{$+$0.112 $\downarrow$} & \textcolor{red}{$-$11.2\% $\downarrow$} \\
\grayrow  &  & \llama & \textbf{0.648} & \textcolor{teal}{$-$0.352 $\uparrow$} & \textcolor{teal}{$+$35.2\% $\uparrow$} \\
\grayrow \multirow{-6}{*}{\makecell[l]{\textbf{6-bar}\\Pop:\,6 / Gen:\,20}} & \multirow{-3}{*}{$\checkmark$} & \qwen & \textbf{0.698} & \textcolor{teal}{$-$0.302 $\uparrow$} & \textcolor{teal}{$+$30.2\% $\uparrow$} \\
\midrule
 &  & \gemma & \textbf{3.039} & \textcolor{red}{$+$2.039 $\downarrow$} & \textcolor{red}{$-$203.9\% $\downarrow$} \\
 &  & \llama & \textbf{1.290} & \textcolor{red}{$+$0.290 $\downarrow$} & \textcolor{red}{$-$29.0\% $\downarrow$} \\
 & \multirow{-3}{*}{$\times$} & \qwen & \textbf{2.088} & \textcolor{red}{$+$1.088 $\downarrow$} & \textcolor{red}{$-$108.8\% $\downarrow$} \\
\cmidrule{2-6}
 &  & \gemma & \textbf{3.092} & \textcolor{red}{$+$2.092 $\downarrow$} & \textcolor{red}{$-$209.2\% $\downarrow$} \\
 &  & \llama & \textbf{1.367} & \textcolor{red}{$+$0.367 $\downarrow$} & \textcolor{red}{$-$36.7\% $\downarrow$} \\
\multirow{-6}{*}{\makecell[l]{\textbf{6-bar}\\Pop:\,60 / Gen:\,300}} & \multirow{-3}{*}{$\checkmark$} & \qwen & \textbf{1.420} & \textcolor{red}{$+$0.420 $\downarrow$} & \textcolor{red}{$-$42.0\% $\downarrow$} \\
\bottomrule
\end{tabular}
\caption{Comparison of models vs.\ GA baselines (Feedback: True). SFB~$\checkmark$ = symbolic feedback enabled; $\times$ = disabled. $\Delta$ and Imp.\,\% are relative to the corresponding GA baseline. \textcolor{teal}{Teal} = improvement over baseline; \textcolor{red}{red} = degradation.}
\label{tab:model_vs_ga_full}
\end{table*}

\subsection{Enum+GA Convergence}
\label{app:ga_convergence}
\begin{figure*}[t]
    \centering
    \includegraphics[width=\textwidth]{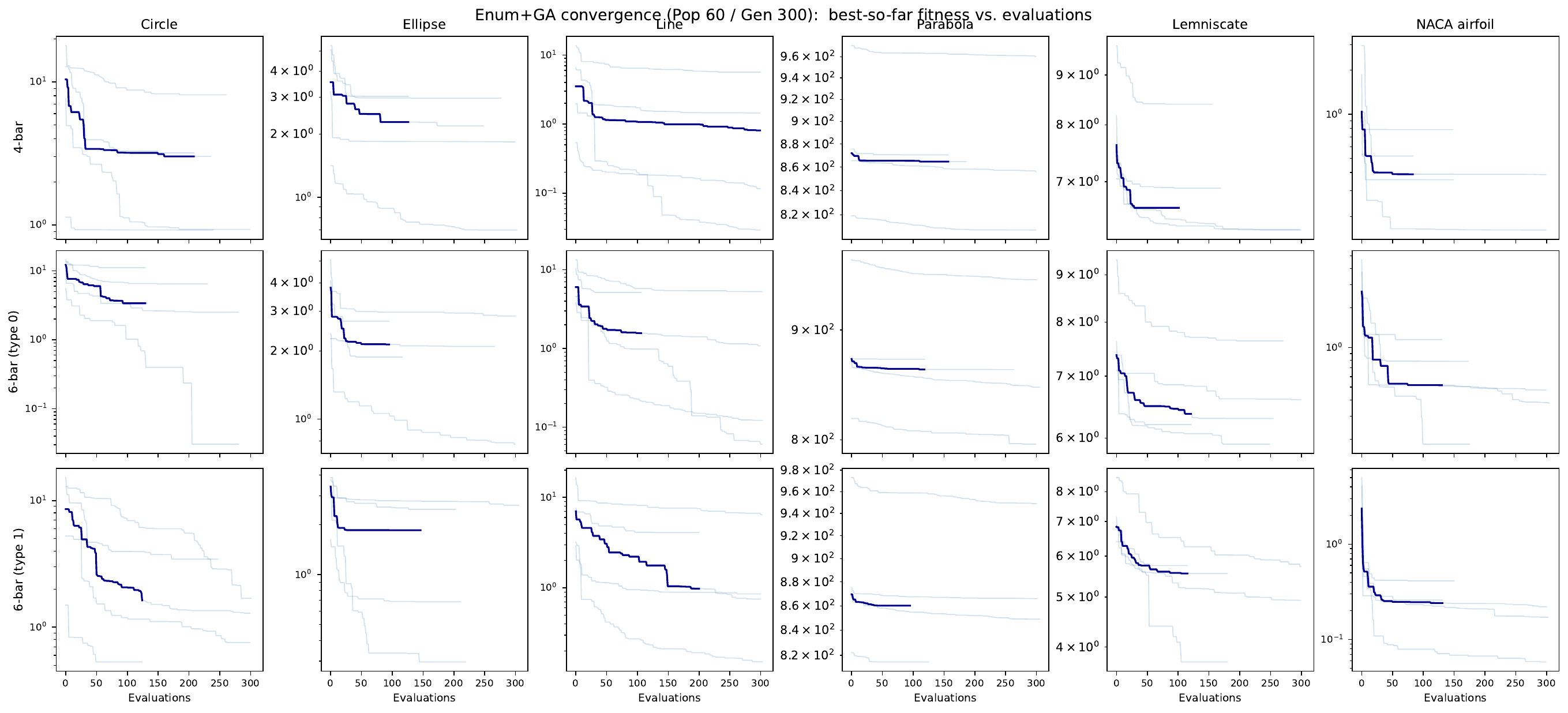}
    \caption{Enum+GA convergence curves (Pop\,60~/ Gen\,300). Best-so-far Chamfer distance (log scale) vs.\ number of evaluations for three linkage topologies (rows) and six shape classes (columns). Thin blue lines: individual runs; bold line: median across 5 runs. All conditions plateau, confirming GA convergence.}
    \label{fig:ga_convergence}
\end{figure*}

Fig.~\ref{fig:ga_convergence} shows the best-so-far fitness (Chamfer distance, log scale) as a function of GA evaluations for the strong Enum+GA baseline (Pop\,60~/ Gen\,300). Each column is a shape class; each row is a linkage topology. Thin lines show individual runs (5 per condition) and the bold line shows the median. In all 18 conditions the curves plateau well before the 18{,}000-evaluation budget is exhausted, confirming that the GA has converged and that the resulting Chamfer distances represent near-optimal solutions for each topology--shape pair.

\subsection{Per-Experiment Metrics}
Table~\ref{tab:all} displays quantitative metrics, such as the average Chamfer distance and the number of convergence steps, for agent-generated outputs.

{\setlength{\tabcolsep}{3pt} 
\begin{table*}
\centering
\resizebox{\textwidth}{!}{\begin{tabular}{@{}llcccccccccc@{}}
\hline
Model & Shape & \#Ex & Fdbk & SFB & Mem & Best chamf. & Fcham & Steps & Fstep & \% Imp. & \% Semantic \\
\hline
	\ac{gemma} & \ac{lb} & 2 & No & No & 0 & $8.847 \pm 0.493$ & $9.078 \pm 0.720$ & $7.333 \pm 2.963$ & $7.000 \pm 3.215$ & $2.203$ & $0.101$ \\
	 & \ac{lb} & 2 & No & No & 2 & $8.893 \pm 0.851$ & $8.893 \pm 0.851$ & $9.000 \pm 4.163$ & $5.000 \pm 3.055$ & $0.000$ & $0.082$ \\
	 & \ac{lb} & 2 & No & Yes & 2 & $8.474 \pm 1.036$ & $8.474 \pm 1.036$ & $9.667 \pm 4.055$ & $9.667 \pm 4.055$ & $0.000$ & $0.168$ \\
	 & \ac{lb} & 2 & Yes & No & 0 & $8.700 \pm 0.375$ & $9.043 \pm 0.395$ & $11.000 \pm 2.608$ & $6.400 \pm 1.600$ & $3.731$ & $0.076$ \\
	 & \ac{lb} & 2 & Yes & No & 2 & $9.132 \pm 0.853$ & $9.965 \pm 0.772$ & $14.400 \pm 1.720$ & $9.600 \pm 2.315$ & $8.171$ & $0.423$ \\
	 & \ac{lb} & 2 & Yes & Yes & 2 & $7.982 \pm 0.729$ & $9.374 \pm 0.444$ & $18.400 \pm 0.245$ & $6.200 \pm 1.497$ & $15.422$ & $0.492$ \\
	 & \ac{lb} & 3 & No & No & 0 & $10.767 \pm 0.985$ & $11.951 \pm 1.432$ & $15.000 \pm 0.577$ & $11.000 \pm 4.509$ & $8.424$ & $0.044$ \\
	 & \ac{lb} & 3 & No & No & 2 & $10.529 \pm 0.675$ & $10.529 \pm 0.675$ & $5.600 \pm 1.720$ & $5.600 \pm 1.720$ & $0.000$ & $0.103$ \\
	 & \ac{lb} & 3 & No & Yes & 2 & $13.662 \pm 1.679$ & $14.552 \pm 1.257$ & $10.333 \pm 4.807$ & $5.667 \pm 3.712$ & $6.332$ & $0.143$ \\
	 & \ac{lb} & 3 & Yes & No & 0 & $8.916 \pm 0.327$ & $10.266 \pm 1.146$ & $8.200 \pm 1.744$ & $4.400 \pm 1.833$ & $10.763$ & $0.079$ \\
	 & \ac{lb} & 3 & Yes & No & 2 & $9.236 \pm 0.561$ & $9.612 \pm 0.533$ & $15.400 \pm 1.691$ & $6.600 \pm 3.265$ & $3.883$ & $0.483$ \\
	 & \ac{lb} & 3 & Yes & Yes & 2 & $9.021 \pm 0.656$ & $9.763 \pm 0.593$ & $16.200 \pm 2.577$ & $6.000 \pm 3.082$ & $7.682$ & $0.545$ \\
     \addlinespace
	 & Circle & 2 & No & No & 0 & $8.490 \pm 2.885$ & $8.503 \pm 2.874$ & $13.750 \pm 2.720$ & $11.500 \pm 2.754$ & $1.059$ & $0.042$ \\
	 & Circle & 2 & No & No & 2 & $11.302 \pm 3.330$ & $11.356 \pm 3.288$ & $7.800 \pm 2.010$ & $7.400 \pm 2.227$ & $4.829$ & $0.208$ \\
	 & Circle & 2 & No & Yes & 2 & $8.227 \pm 3.201$ & $8.242 \pm 3.189$ & $13.750 \pm 3.276$ & $13.000 \pm 3.109$ & $1.643$ & $0.089$ \\
	 & Circle & 2 & Yes & No & 0 & $11.924 \pm 3.486$ & $14.248 \pm 3.615$ & $10.400 \pm 2.977$ & $5.800 \pm 2.035$ & $14.806$ & $0.053$ \\
	 & Circle & 2 & Yes & No & 2 & $10.899 \pm 3.318$ & $12.040 \pm 3.422$ & $15.200 \pm 1.881$ & $7.600 \pm 1.939$ & $9.651$ & $0.527$ \\
	 & Circle & 2 & Yes & Yes & 2 & $9.919 \pm 3.420$ & $11.597 \pm 3.365$ & $10.600 \pm 2.768$ & $1.600 \pm 0.400$ & $23.521$ & $0.649$ \\
	 & Circle & 3 & No & No & 0 & $11.490 \pm 2.669$ & $11.509 \pm 2.654$ & $6.000 \pm 1.225$ & $4.200 \pm 1.068$ & $0.611$ & $0.050$ \\
	 & Circle & 3 & No & No & 2 & $9.019 \pm 2.241$ & $11.203 \pm 3.348$ & $11.400 \pm 2.768$ & $11.000 \pm 2.950$ & $10.285$ & $0.162$ \\
	 & Circle & 3 & No & Yes & 2 & $6.288 \pm 1.658$ & $8.826 \pm 2.147$ & $13.600 \pm 2.638$ & $10.000 \pm 2.387$ & $20.266$ & $0.187$ \\
	 & Circle & 3 & Yes & No & 0 & $10.074 \pm 2.260$ & $11.837 \pm 3.345$ & $11.600 \pm 2.159$ & $8.000 \pm 3.178$ & $9.397$ & $0.076$ \\
	 & Circle & 3 & Yes & No & 2 & $7.351 \pm 1.974$ & $11.817 \pm 3.059$ & $13.000 \pm 3.050$ & $3.800 \pm 1.356$ & $29.604$ & $0.595$ \\
	 & Circle & 3 & Yes & Yes & 2 & $10.416 \pm 3.413$ & $11.744 \pm 3.908$ & $14.200 \pm 2.289$ & $3.400 \pm 1.030$ & $7.921$ & $0.519$ \\
     \addlinespace
	 & Ellipse & 2 & No & No & 0 & $6.677 \pm 1.335$ & $6.677 \pm 1.335$ & $9.000 \pm 4.726$ & $9.000 \pm 4.726$ & $0.000$ & $0.029$ \\
	 & Ellipse & 2 & No & No & 2 & $6.945 \pm 2.819$ & $6.945 \pm 2.819$ & $11.000 \pm 9.000$ & $11.000 \pm 9.000$ & $0.000$ & $0.026$ \\
	 & Ellipse & 2 & No & Yes & 2 & $6.568 \pm 1.315$ & $6.568 \pm 1.315$ & $3.667 \pm 1.764$ & $3.667 \pm 1.764$ & $0.000$ & $0.176$ \\
	 & Ellipse & 2 & Yes & No & 0 & $7.985 \pm 1.375$ & $8.928 \pm 1.461$ & $10.800 \pm 1.828$ & $7.400 \pm 1.778$ & $10.527$ & $0.039$ \\
	 & Ellipse & 2 & Yes & No & 2 & $6.355 \pm 1.001$ & $6.714 \pm 1.133$ & $11.600 \pm 2.786$ & $8.800 \pm 2.596$ & $4.053$ & $0.442$ \\
	 & Ellipse & 2 & Yes & Yes & 2 & $6.836 \pm 0.991$ & $7.348 \pm 1.079$ & $12.000 \pm 2.429$ & $10.400 \pm 3.234$ & $6.179$ & $0.218$ \\
	 & Ellipse & 3 & No & No & 0 & $6.722 \pm 1.353$ & $8.009 \pm 1.415$ & $6.800 \pm 2.498$ & $4.000 \pm 1.265$ & $17.586$ & $0.211$ \\
	 & Ellipse & 3 & No & No & 2 & $7.678 \pm 1.267$ & $7.678 \pm 1.267$ & $7.800 \pm 2.709$ & $7.800 \pm 2.709$ & $0.000$ & $0.213$ \\
	 & Ellipse & 3 & No & Yes & 2 & $7.372 \pm 1.454$ & $7.659 \pm 1.276$ & $7.600 \pm 2.135$ & $5.600 \pm 2.272$ & $6.428$ & $0.371$ \\
	 & Ellipse & 3 & Yes & No & 0 & $6.900 \pm 1.443$ & $7.777 \pm 1.281$ & $12.600 \pm 2.542$ & $3.800 \pm 2.332$ & $14.609$ & $0.121$ \\
	 & Ellipse & 3 & Yes & No & 2 & $6.647 \pm 1.244$ & $7.445 \pm 1.396$ & $13.600 \pm 3.027$ & $6.400 \pm 2.713$ & $11.320$ & $0.472$ \\
	 & Ellipse & 3 & Yes & Yes & 2 & $5.552 \pm 0.819$ & $6.665 \pm 1.103$ & $12.000 \pm 4.324$ & $2.000 \pm 0.775$ & $15.391$ & $0.394$ \\
\end{tabular}}
\caption*{}
\phantomcaption
\label{tab:all}
\end{table*}

\begin{table*}
\renewcommand\thetable{3}
\centering
\centering
\resizebox{\textwidth}{!}{\begin{tabular}{@{}llcccccccccc@{}}
\hline
Model & Shape & \#Ex & Fdbk & SFB & Mem & Best chamf. & Fcham & Steps & Fstep & \% Imp. & \% Semantic \\
\hline
	 \ac{gemma} & Line & 2 & No & No & 0 & $13.640 \pm 0.322$ & $13.640 \pm 0.322$ & $11.000 \pm 3.000$ & $11.000 \pm 3.000$ & $0.000$ & $0.002$ \\
	 & Line & 2 & No & No & 2 & $13.643 \pm 1.385$ & $13.643 \pm 1.385$ & $10.667 \pm 4.096$ & $10.667 \pm 4.096$ & $0.000$ & $0.136$ \\
	 & Line & 2 & No & Yes & 2 & $12.076 \pm 0.947$ & $14.277 \pm 1.254$ & $8.500 \pm 3.500$ & $7.500 \pm 2.500$ & $14.172$ & $0.778$ \\
	 & Line & 2 & Yes & No & 0 & $11.150 \pm 1.542$ & $11.664 \pm 1.139$ & $10.400 \pm 3.234$ & $9.800 \pm 3.484$ & $6.050$ & $0.032$ \\
	 & Line & 2 & Yes & No & 2 & $10.827 \pm 1.545$ & $11.298 \pm 1.214$ & $12.200 \pm 1.530$ & $8.800 \pm 2.059$ & $6.025$ & $0.241$ \\
	 & Line & 2 & Yes & Yes & 2 & $10.186 \pm 2.444$ & $10.186 \pm 2.444$ & $10.600 \pm 2.750$ & $10.000 \pm 2.449$ & $0.000$ & $0.151$ \\
	 & Line & 3 & No & No & 0 & $8.619 \pm 3.041$ & $9.501 \pm 3.392$ & $6.750 \pm 4.423$ & $6.500 \pm 4.518$ & $6.019$ & $0.038$ \\
	 & Line & 3 & No & No & 2 & $11.917 \pm 1.188$ & $11.917 \pm 1.188$ & $9.000 \pm 3.055$ & $9.000 \pm 3.055$ & $0.000$ & $0.221$ \\
	 & Line & 3 & No & Yes & 2 & $12.071 \pm 0.942$ & $12.071 \pm 0.942$ & $4.500 \pm 1.500$ & $4.500 \pm 1.500$ & $0.000$ & $0.319$ \\
	 & Line & 3 & Yes & No & 0 & $9.715 \pm 1.258$ & $10.689 \pm 0.921$ & $6.800 \pm 1.393$ & $3.600 \pm 1.208$ & $10.022$ & $0.056$ \\
	 & Line & 3 & Yes & No & 2 & $10.043 \pm 1.210$ & $10.835 \pm 1.468$ & $10.200 \pm 2.518$ & $6.000 \pm 2.387$ & $6.166$ & $0.552$ \\
	 & Line & 3 & Yes & Yes & 2 & $11.035 \pm 0.986$ & $12.417 \pm 1.243$ & $14.800 \pm 2.939$ & $5.600 \pm 1.806$ & $10.570$ & $0.371$ \\
    \addlinespace

	 & Naca & 2 & No & No & 0 & $1.017 \pm 0.436$ & $1.017 \pm 0.436$ & $7.000 \pm 1.000$ & $7.000 \pm 1.000$ & $0.000$ & $0.058$ \\
	 & Naca & 2 & No & No & 2 & $1.104 \pm 0.151$ & $1.104 \pm 0.151$ & $6.333 \pm 0.882$ & $6.333 \pm 0.882$ & $0.000$ & $0.041$ \\
	 & Naca & 2 & No & Yes & 2 & $1.057 \pm 0.416$ & $1.057 \pm 0.416$ & $7.000 \pm 0.000$ & $7.000 \pm 0.000$ & $0.000$ & $0.026$ \\
	 & Naca & 2 & Yes & No & 0 & $1.082 \pm 0.416$ & $1.336 \pm 0.306$ & $12.600 \pm 2.040$ & $7.800 \pm 1.715$ & $27.086$ & $0.078$ \\
	 & Naca & 2 & Yes & No & 2 & $1.618 \pm 0.555$ & $1.714 \pm 0.508$ & $16.400 \pm 1.364$ & $11.800 \pm 3.611$ & $11.084$ & $0.313$ \\
	 & Naca & 2 & Yes & Yes & 2 & $1.074 \pm 0.371$ & $1.663 \pm 0.523$ & $12.000 \pm 2.470$ & $8.200 \pm 2.709$ & $30.463$ & $0.522$ \\
	 & Naca & 3 & No & No & 0 & $1.498 \pm 0.499$ & $1.628 \pm 0.427$ & $7.600 \pm 2.694$ & $4.600 \pm 1.913$ & $15.066$ & $0.159$ \\
	 & Naca & 3 & No & No & 2 & $1.245 \pm 0.533$ & $1.245 \pm 0.533$ & $7.000 \pm 3.507$ & $7.000 \pm 3.507$ & $0.000$ & $0.209$ \\
	 & Naca & 3 & No & Yes & 2 & $1.687 \pm 0.518$ & $1.883 \pm 0.544$ & $8.000 \pm 1.975$ & $5.200 \pm 1.685$ & $18.053$ & $0.328$ \\
	 & Naca & 3 & Yes & No & 0 & $1.347 \pm 0.404$ & $2.381 \pm 0.462$ & $6.600 \pm 1.400$ & $3.200 \pm 1.428$ & $46.926$ & $0.146$ \\
	 & Naca & 3 & Yes & No & 2 & $0.997 \pm 0.307$ & $2.021 \pm 0.512$ & $18.200 \pm 0.860$ & $5.000 \pm 1.897$ & $49.954$ & $0.566$ \\
	 & Naca & 3 & Yes & Yes & 2 & $0.796 \pm 0.204$ & $1.702 \pm 0.439$ & $13.800 \pm 2.458$ & $2.000 \pm 0.548$ & $42.740$ & $0.854$ \\
		\addlinespace
	 & Parabola & 2 & No & No & 0 & $869.205 \pm 35.569$ & $869.205 \pm 35.569$ & $9.500 \pm 1.500$ & $9.500 \pm 1.500$ & $0.000$ & $0.036$ \\
	 & Parabola & 2 & No & No & 2 & $861.442 \pm 28.839$ & $861.442 \pm 28.839$ & $6.000 \pm 4.000$ & $6.000 \pm 4.000$ & $0.000$ & $0.352$ \\
	 & Parabola & 2 & No & Yes & 2 & $906.927 \pm 50.681$ & $906.927 \pm 50.681$ & $10.667 \pm 3.712$ & $10.667 \pm 3.712$ & $0.000$ & $0.019$ \\
	 & Parabola & 2 & Yes & No & 0 & $890.160 \pm 23.248$ & $890.612 \pm 23.162$ & $13.800 \pm 3.338$ & $10.800 \pm 3.513$ & $0.052$ & $0.023$ \\
	 & Parabola & 2 & Yes & No & 2 & $892.928 \pm 23.123$ & $897.871 \pm 27.055$ & $15.000 \pm 2.168$ & $6.000 \pm 2.145$ & $0.502$ & $0.304$ \\
	 & Parabola & 2 & Yes & Yes & 2 & $858.509 \pm 40.475$ & $892.209 \pm 23.039$ & $12.600 \pm 2.159$ & $8.200 \pm 1.744$ & $3.784$ & $0.300$ \\
	 & Parabola & 3 & No & No & 0 & $880.197 \pm 16.446$ & $881.267 \pm 16.420$ & $8.600 \pm 1.631$ & $5.200 \pm 0.860$ & $0.121$ & $0.073$ \\
	 & Parabola & 3 & No & No & 2 & $897.018 \pm 33.711$ & $898.569 \pm 33.387$ & $7.600 \pm 2.839$ & $5.200 \pm 2.223$ & $0.177$ & $0.505$ \\
	 & Parabola & 3 & No & Yes & 2 & $871.551 \pm 20.077$ & $871.551 \pm 20.077$ & $5.000 \pm 3.050$ & $5.000 \pm 3.050$ & $0.000$ & $0.497$ \\
	 & Parabola & 3 & Yes & No & 0 & $889.653 \pm 22.191$ & $896.919 \pm 24.107$ & $11.600 \pm 2.768$ & $6.400 \pm 1.288$ & $0.788$ & $0.072$ \\
	 & Parabola & 3 & Yes & No & 2 & $895.262 \pm 24.071$ & $905.689 \pm 33.476$ & $10.200 \pm 2.818$ & $7.200 \pm 2.746$ & $1.013$ & $0.396$ \\
	 & Parabola & 3 & Yes & Yes & 2 & $892.568 \pm 23.659$ & $894.953 \pm 22.691$ & $10.400 \pm 2.750$ & $5.800 \pm 2.354$ & $0.278$ & $0.513$ \\
\end{tabular}}
\end{table*}

\begin{table*}
\renewcommand\thetable{3}
\centering
\centering
\resizebox{\textwidth}{!}{\begin{tabular}{@{}llcccccccccc@{}}
\hline
Model & Shape & \#Ex & Fdbk & SFB & Mem & Best chamf. & Fcham & Steps & Fstep & \% Imp. & \% Semantic \\
\hline
	\ac{llama} & \ac{lb} & 2 & No & No & 0 & $7.415 \pm 0.864$ & $14.752 \pm 2.231$ & $14.750 \pm 2.136$ & $1.500 \pm 0.289$ & $45.126$ & $0.231$ \\
	 & \ac{lb} & 2 & No & No & 2 & $7.730 \pm 0.445$ & $10.356 \pm 1.944$ & $3.600 \pm 1.631$ & $1.400 \pm 0.245$ & $18.120$ & $0.786$ \\
	 & \ac{lb} & 2 & No & Yes & 2 & $8.264 \pm 0.759$ & $11.495 \pm 1.094$ & $5.800 \pm 2.615$ & $1.800 \pm 0.490$ & $25.670$ & $0.816$ \\
	 & \ac{lb} & 2 & Yes & No & 0 & $7.029 \pm 0.825$ & $12.083 \pm 1.603$ & $10.400 \pm 3.219$ & $1.000 \pm 0.000$ & $35.867$ & $0.375$ \\
	 & \ac{lb} & 2 & Yes & No & 2 & $7.539 \pm 0.638$ & $12.458 \pm 2.201$ & $9.250 \pm 2.810$ & $1.750 \pm 0.479$ & $34.418$ & $0.899$ \\
	 & \ac{lb} & 2 & Yes & Yes & 2 & $7.986 \pm 0.701$ & $16.445 \pm 2.855$ & $11.000 \pm 3.416$ & $1.250 \pm 0.250$ & $45.705$ & $0.801$ \\
	 & \ac{lb} & 3 & No & No & 0 & $7.861 \pm 0.522$ & $15.193 \pm 2.333$ & $10.600 \pm 3.709$ & $1.400 \pm 0.245$ & $43.505$ & $0.324$ \\
	 & \ac{lb} & 3 & No & No & 2 & $10.945 \pm 1.811$ & $12.727 \pm 3.415$ & $3.750 \pm 2.136$ & $1.250 \pm 0.250$ & $8.388$ & $0.796$ \\
	 & \ac{lb} & 3 & No & Yes & 2 & $11.656 \pm 1.987$ & $16.404 \pm 2.378$ & $4.600 \pm 1.806$ & $1.800 \pm 0.583$ & $25.118$ & $0.576$ \\
	 & \ac{lb} & 3 & Yes & No & 0 & $6.610 \pm 1.389$ & $17.877 \pm 3.803$ & $14.667 \pm 2.728$ & $1.000 \pm 0.000$ & $57.215$ & $0.259$ \\
	 & \ac{lb} & 3 & Yes & No & 2 & $8.017 \pm 1.204$ & $10.287 \pm 1.723$ & $4.500 \pm 3.500$ & $1.000 \pm 0.000$ & $21.837$ & $0.969$ \\
	 & \ac{lb} & 3 & Yes & Yes & 2 & $7.337 \pm 0.212$ & $10.209 \pm 1.356$ & $17.333 \pm 1.764$ & $1.000 \pm 0.000$ & $25.852$ & $0.855$ \\
		\addlinespace
	 & Circle & 2 & No & No & 0 & $2.673 \pm 0.714$ & $12.537 \pm 3.937$ & $8.800 \pm 3.023$ & $2.200 \pm 0.583$ & $77.134$ & $0.192$ \\
	 & Circle & 2 & No & No & 2 & $11.047 \pm 3.018$ & $11.119 \pm 2.962$ & $4.000 \pm 1.291$ & $3.500 \pm 1.500$ & $1.730$ & $0.550$ \\
	 & Circle & 2 & No & Yes & 2 & $7.901 \pm 3.435$ & $11.478 \pm 2.628$ & $14.200 \pm 2.518$ & $6.200 \pm 1.855$ & $40.320$ & $0.632$ \\
	 & Circle & 2 & Yes & No & 0 & $1.466 \pm 0.659$ & $12.239 \pm 3.669$ & $17.000 \pm 1.225$ & $1.500 \pm 0.289$ & $90.609$ & $0.238$ \\
	 & Circle & 2 & Yes & No & 2 & $5.482 \pm 2.233$ & $12.932 \pm 4.846$ & $11.500 \pm 3.069$ & $1.750 \pm 0.750$ & $52.205$ & $0.758$ \\
	 & Circle & 2 & Yes & Yes & 2 & $1.930 \pm 0.428$ & $8.044 \pm 3.258$ & $17.500 \pm 0.645$ & $4.750 \pm 3.119$ & $58.285$ & $0.676$ \\
	 & Circle & 3 & No & No & 0 & $4.011 \pm 2.143$ & $6.372 \pm 2.802$ & $7.400 \pm 3.385$ & $2.000 \pm 0.447$ & $29.322$ & $0.170$ \\
	 & Circle & 3 & No & No & 2 & $7.237 \pm 6.154$ & $9.361 \pm 7.705$ & $9.667 \pm 3.712$ & $4.000 \pm 3.000$ & $22.474$ & $0.723$ \\
	 & Circle & 3 & No & Yes & 2 & $6.334 \pm 3.305$ & $11.281 \pm 3.148$ & $11.400 \pm 2.159$ & $2.400 \pm 0.678$ & $54.314$ & $0.646$ \\
	 & Circle & 3 & Yes & No & 0 & $2.181 \pm 0.848$ & $10.715 \pm 4.386$ & $7.200 \pm 2.223$ & $1.200 \pm 0.200$ & $73.424$ & $0.387$ \\
	 & Circle & 3 & Yes & No & 2 & $4.097 \pm 2.085$ & $11.403 \pm 3.930$ & $18.200 \pm 1.562$ & $1.400 \pm 0.400$ & $60.651$ & $0.702$ \\
	 & Circle & 3 & Yes & Yes & 2 & $5.660 \pm 2.889$ & $10.780 \pm 3.395$ & $11.600 \pm 2.249$ & $1.800 \pm 0.800$ & $54.820$ & $0.826$ \\
		\addlinespace
	 & Ellipse & 2 & No & No & 0 & $4.218 \pm 0.505$ & $7.903 \pm 2.433$ & $9.600 \pm 2.542$ & $1.600 \pm 0.400$ & $30.905$ & $0.164$ \\
	 & Ellipse & 2 & No & No & 2 & $4.523 \pm 0.673$ & $7.153 \pm 1.232$ & $10.200 \pm 2.672$ & $2.000 \pm 0.316$ & $34.718$ & $0.837$ \\
	 & Ellipse & 2 & No & Yes & 2 & $4.599 \pm 0.977$ & $6.798 \pm 1.167$ & $10.200 \pm 1.828$ & $2.400 \pm 0.510$ & $30.822$ & $0.731$ \\
	 & Ellipse & 2 & Yes & No & 0 & $4.113 \pm 0.590$ & $8.028 \pm 1.123$ & $14.400 \pm 1.600$ & $1.800 \pm 0.374$ & $45.818$ & $0.252$ \\
	 & Ellipse & 2 & Yes & No & 2 & $3.383 \pm 0.391$ & $6.356 \pm 0.877$ & $17.000 \pm 1.732$ & $2.750 \pm 1.436$ & $43.673$ & $0.862$ \\
	 & Ellipse & 2 & Yes & Yes & 2 & $3.371 \pm 0.906$ & $10.413 \pm 5.775$ & $12.000 \pm 1.826$ & $2.750 \pm 1.750$ & $52.252$ & $0.737$ \\
	 & Ellipse & 3 & No & No & 0 & $3.801 \pm 0.663$ & $8.619 \pm 1.584$ & $9.600 \pm 2.713$ & $2.000 \pm 0.548$ & $52.293$ & $0.217$ \\
	 & Ellipse & 3 & No & No & 2 & $5.523 \pm 0.598$ & $11.445 \pm 2.579$ & $3.800 \pm 0.860$ & $2.000 \pm 0.632$ & $43.739$ & $0.826$ \\
	 & Ellipse & 3 & No & Yes & 2 & $4.759 \pm 0.820$ & $7.138 \pm 0.904$ & $15.200 \pm 2.267$ & $2.600 \pm 1.166$ & $31.002$ & $0.633$ \\
	 & Ellipse & 3 & Yes & No & 0 & $4.048 \pm 0.798$ & $7.090 \pm 1.704$ & $4.000 \pm 1.683$ & $1.250 \pm 0.250$ & $41.283$ & $0.303$ \\
	 & Ellipse & 3 & Yes & No & 2 & $2.935 \pm 0.621$ & $8.870 \pm 4.465$ & $11.333 \pm 1.202$ & $2.333 \pm 0.882$ & $54.619$ & $0.606$ \\
	 & Ellipse & 3 & Yes & Yes & 2 & $3.429 \pm 0.496$ & $7.536 \pm 2.523$ & $8.667 \pm 4.702$ & $2.000 \pm 0.577$ & $41.305$ & $0.785$ \\
\end{tabular}}
\end{table*}

\begin{table*}
\renewcommand\thetable{3}
\centering
\centering
\resizebox{\textwidth}{!}{\begin{tabular}{@{}llcccccccccc@{}}
\hline
Model & Shape & \#Ex & Fdbk & SFB & Mem & Best chamf. & Fcham & Steps & Fstep & \% Imp. & \% Semantic \\
\hline
	\ac{llama} & Line & 2 & No & No & 0 & $3.479 \pm 1.495$ & $10.273 \pm 1.666$ & $12.400 \pm 1.939$ & $3.000 \pm 0.894$ & $65.454$ & $0.113$ \\
	 & Line & 2 & No & No & 2 & $5.999 \pm 1.191$ & $9.055 \pm 1.468$ & $9.200 \pm 3.023$ & $4.800 \pm 2.577$ & $30.440$ & $0.541$ \\
	 & Line & 2 & No & Yes & 2 & $5.089 \pm 1.962$ & $6.117 \pm 1.809$ & $11.000 \pm 2.074$ & $7.000 \pm 2.121$ & $18.628$ & $0.536$ \\
	 & Line & 2 & Yes & No & 0 & $2.935 \pm 1.658$ & $8.540 \pm 2.595$ & $8.000 \pm 1.581$ & $1.750 \pm 0.750$ & $66.011$ & $0.138$ \\
	 & Line & 2 & Yes & No & 2 & $2.024 \pm 0.803$ & $7.411 \pm 1.205$ & $9.750 \pm 3.660$ & $1.750 \pm 0.750$ & $62.358$ & $0.653$ \\
	 & Line & 2 & Yes & Yes & 2 & $4.203 \pm 2.095$ & $11.977 \pm 1.247$ & $16.000 \pm 1.761$ & $1.200 \pm 0.200$ & $66.938$ & $0.930$ \\
	 & Line & 3 & No & No & 0 & $1.892 \pm 1.151$ & $6.488 \pm 1.696$ & $13.000 \pm 3.000$ & $1.600 \pm 0.400$ & $70.869$ & $0.167$ \\
	 & Line & 3 & No & No & 2 & $2.824 \pm 1.109$ & $7.774 \pm 2.180$ & $5.000 \pm 1.080$ & $2.000 \pm 0.408$ & $47.051$ & $0.794$ \\
	 & Line & 3 & No & Yes & 2 & $6.741 \pm 2.275$ & $9.788 \pm 1.316$ & $9.000 \pm 1.871$ & $2.600 \pm 0.980$ & $35.589$ & $0.661$ \\
	 & Line & 3 & Yes & No & 0 & $0.904 \pm 0.402$ & $11.338 \pm 3.167$ & $7.250 \pm 1.109$ & $1.000 \pm 0.000$ & $89.959$ & $0.178$ \\
	 & Line & 3 & Yes & No & 2 & $3.281 \pm 1.334$ & $9.965 \pm 3.968$ & $14.000 \pm 1.155$ & $1.667 \pm 0.333$ & $51.229$ & $0.675$ \\
	 & Line & 3 & Yes & Yes & 2 & $3.090 \pm 1.299$ & $10.357 \pm 2.300$ & $14.000 \pm 2.588$ & $2.400 \pm 0.600$ & $68.001$ & $0.758$ \\
		\addlinespace

	 & Naca & 2 & No & No & 0 & $1.555 \pm 0.635$ & $2.577 \pm 0.494$ & $11.600 \pm 3.043$ & $5.400 \pm 3.655$ & $38.687$ & $0.157$ \\
	 & Naca & 2 & No & No & 2 & $1.247 \pm 0.444$ & $2.166 \pm 0.604$ & $4.000 \pm 1.049$ & $2.800 \pm 1.114$ & $46.449$ & $0.624$ \\
	 & Naca & 2 & No & Yes & 2 & $1.597 \pm 0.532$ & $1.714 \pm 0.472$ & $5.600 \pm 3.187$ & $4.800 \pm 3.323$ & $12.779$ & $0.588$ \\
	 & Naca & 2 & Yes & No & 0 & $0.624 \pm 0.182$ & $1.722 \pm 0.494$ & $10.400 \pm 3.763$ & $1.600 \pm 0.400$ & $57.032$ & $0.309$ \\
	 & Naca & 2 & Yes & No & 2 & $0.655 \pm 0.209$ & $2.469 \pm 0.303$ & $13.750 \pm 1.652$ & $1.500 \pm 0.289$ & $70.067$ & $0.873$ \\
	 & Naca & 2 & Yes & Yes & 2 & $0.922 \pm 0.313$ & $1.614 \pm 0.701$ & $9.667 \pm 5.175$ & $1.000 \pm 0.000$ & $29.433$ & $0.888$ \\
	 & Naca & 3 & No & No & 0 & $1.157 \pm 0.373$ & $2.021 \pm 0.794$ & $9.400 \pm 2.542$ & $1.400 \pm 0.245$ & $31.172$ & $0.194$ \\
	 & Naca & 3 & No & No & 2 & $1.315 \pm 0.367$ & $1.643 \pm 0.630$ & $4.250 \pm 2.016$ & $2.000 \pm 0.707$ & $11.631$ & $0.828$ \\
	 & Naca & 3 & No & Yes & 2 & $1.168 \pm 0.367$ & $2.324 \pm 0.740$ & $10.600 \pm 2.337$ & $4.600 \pm 2.619$ & $43.976$ & $0.564$ \\
	 & Naca & 3 & Yes & No & 0 & $0.943 \pm 0.419$ & $2.576 \pm 0.817$ & $5.750 \pm 2.496$ & $1.750 \pm 0.479$ & $66.220$ & $0.381$ \\
	 & Naca & 3 & Yes & No & 2 & $1.456 \pm 0.571$ & $2.291 \pm 0.235$ & $10.333 \pm 4.631$ & $2.000 \pm 1.000$ & $38.924$ & $0.675$ \\
	 & Naca & 3 & Yes & Yes & 2 & $1.016 \pm 0.427$ & $1.939 \pm 0.472$ & $17.500 \pm 1.658$ & $1.250 \pm 0.250$ & $49.899$ & $0.776$ \\
		\addlinespace
	 & Parabola & 2 & No & No & 0 & $871.702 \pm 29.579$ & $892.003 \pm 24.567$ & $13.400 \pm 3.234$ & $2.000 \pm 0.548$ & $2.319$ & $0.224$ \\
	 & Parabola & 2 & No & No & 2 & $872.853 \pm 34.498$ & $894.611 \pm 30.983$ & $9.800 \pm 3.625$ & $2.600 \pm 0.748$ & $2.478$ & $0.748$ \\
	 & Parabola & 2 & No & Yes & 2 & $851.307 \pm 33.603$ & $909.159 \pm 20.136$ & $8.000 \pm 2.588$ & $3.400 \pm 2.159$ & $6.395$ & $0.775$ \\
	 & Parabola & 2 & Yes & No & 0 & $866.615 \pm 32.053$ & $1125.509 \pm 211.739$ & $11.600 \pm 2.379$ & $1.400 \pm 0.245$ & $14.196$ & $0.302$ \\
	 & Parabola & 2 & Yes & No & 2 & $817.244 \pm 83.559$ & $900.522 \pm 23.241$ & $12.000 \pm 2.775$ & $1.200 \pm 0.200$ & $9.727$ & $0.827$ \\
	 & Parabola & 2 & Yes & Yes & 2 & $815.179 \pm 84.612$ & $896.679 \pm 26.440$ & $10.200 \pm 2.764$ & $1.200 \pm 0.200$ & $9.732$ & $0.836$ \\
	 & Parabola & 3 & No & No & 0 & $838.519 \pm 33.608$ & $879.870 \pm 18.275$ & $16.800 \pm 1.985$ & $3.200 \pm 0.583$ & $4.761$ & $0.204$ \\
	 & Parabola & 3 & No & No & 2 & $845.166 \pm 45.083$ & $858.242 \pm 41.509$ & $5.600 \pm 1.122$ & $2.000 \pm 0.775$ & $1.619$ & $0.731$ \\
	 & Parabola & 3 & No & Yes & 2 & $843.717 \pm 36.534$ & $902.770 \pm 49.407$ & $11.600 \pm 2.977$ & $1.200 \pm 0.200$ & $5.900$ & $0.787$ \\
	 & Parabola & 3 & Yes & No & 0 & $838.254 \pm 37.670$ & $896.905 \pm 23.839$ & $14.400 \pm 2.064$ & $1.600 \pm 0.400$ & $6.597$ & $0.294$ \\
	 & Parabola & 3 & Yes & No & 2 & $843.277 \pm 49.965$ & $898.882 \pm 26.610$ & $11.200 \pm 2.223$ & $1.000 \pm 0.000$ & $6.205$ & $0.820$ \\
	 & Parabola & 3 & Yes & Yes & 2 & $848.027 \pm 35.591$ & $898.265 \pm 24.218$ & $16.600 \pm 0.678$ & $1.000 \pm 0.000$ & $5.670$ & $0.872$ \\
		\addlinespace        
\end{tabular}}
\end{table*}

\begin{table*}
\renewcommand\thetable{3}
\centering
\centering
\resizebox{\textwidth}{!}{\begin{tabular}{@{}llcccccccccc@{}}
\hline
Model & Shape & \#Ex & Fdbk & SFB & Mem & Best chamf. & Fcham & Steps & Fstep & \% Imp. & \% Semantic \\
\hline
	\ac{qwen} & \ac{lb} & 2 & No & No & 0 & $8.589 \pm 0.319$ & $10.122 \pm 0.512$ & $7.400 \pm 3.234$ & $1.800 \pm 0.374$ & $13.760$ & $0.367$ \\
	 & \ac{lb} & 2 & No & No & 2 & $8.530 \pm 0.376$ & $12.677 \pm 1.963$ & $9.600 \pm 3.473$ & $1.000 \pm 0.000$ & $26.693$ & $0.512$ \\
	 & \ac{lb} & 2 & No & Yes & 2 & $9.129 \pm 0.866$ & $10.727 \pm 1.141$ & $11.200 \pm 2.888$ & $1.800 \pm 0.374$ & $13.545$ & $0.611$ \\
	 & \ac{lb} & 2 & Yes & No & 0 & $8.047 \pm 0.683$ & $10.937 \pm 1.311$ & $13.200 \pm 1.855$ & $1.400 \pm 0.245$ & $25.157$ & $0.288$ \\
	 & \ac{lb} & 2 & Yes & No & 2 & $8.763 \pm 0.943$ & $11.410 \pm 1.425$ & $6.800 \pm 2.396$ & $1.200 \pm 0.200$ & $22.070$ & $0.471$ \\
	 & \ac{lb} & 2 & Yes & Yes & 2 & $7.830 \pm 1.112$ & $10.027 \pm 0.589$ & $14.000 \pm 2.214$ & $1.800 \pm 0.374$ & $23.076$ & $0.467$ \\
	 & \ac{lb} & 3 & No & No & 0 & $8.460 \pm 0.586$ & $9.789 \pm 0.486$ & $11.400 \pm 3.156$ & $1.200 \pm 0.200$ & $13.869$ & $0.381$ \\
	 & \ac{lb} & 3 & No & No & 2 & $8.075 \pm 0.656$ & $12.422 \pm 2.085$ & $15.200 \pm 2.888$ & $1.200 \pm 0.200$ & $27.637$ & $0.349$ \\
	 & \ac{lb} & 3 & No & Yes & 2 & $7.832 \pm 0.183$ & $9.717 \pm 0.465$ & $12.000 \pm 2.588$ & $1.600 \pm 0.400$ & $18.843$ & $0.467$ \\
	 & \ac{lb} & 3 & Yes & No & 0 & $7.291 \pm 0.632$ & $10.550 \pm 1.101$ & $10.400 \pm 1.122$ & $1.000 \pm 0.000$ & $28.501$ & $0.310$ \\
	 & \ac{lb} & 3 & Yes & No & 2 & $6.905 \pm 1.095$ & $9.901 \pm 0.439$ & $11.400 \pm 2.657$ & $1.000 \pm 0.000$ & $28.431$ & $0.504$ \\
	 & \ac{lb} & 3 & Yes & Yes & 2 & $7.804 \pm 0.396$ & $10.686 \pm 0.758$ & $14.400 \pm 2.337$ & $1.000 \pm 0.000$ & $25.554$ & $0.638$ \\
		\addlinespace
	 & Circle & 2 & No & No & 0 & $11.664 \pm 3.720$ & $11.664 \pm 3.720$ & $9.667 \pm 3.844$ & $9.667 \pm 3.844$ & $0.000$ & $0.019$ \\
	 & Circle & 2 & No & No & 2 & $12.950 \pm 2.919$ & $12.950 \pm 2.919$ & $12.500 \pm 3.379$ & $12.500 \pm 3.379$ & $0.000$ & $0.019$ \\
	 & Circle & 2 & No & Yes & 2 & $11.226 \pm 3.332$ & $11.226 \pm 3.332$ & $14.333 \pm 0.333$ & $14.333 \pm 0.333$ & $0.000$ & $0.020$ \\
	 & Circle & 2 & Yes & No & 0 & $10.478 \pm 4.392$ & $10.478 \pm 4.392$ & $11.200 \pm 1.855$ & $11.200 \pm 1.855$ & $0.000$ & $0.011$ \\
	 & Circle & 2 & Yes & No & 2 & $3.139 \pm 1.309$ & $6.518 \pm 2.842$ & $7.600 \pm 2.462$ & $7.400 \pm 2.337$ & $19.826$ & $0.087$ \\
	 & Circle & 2 & Yes & Yes & 2 & $5.880 \pm 5.585$ & $5.880 \pm 5.585$ & $10.667 \pm 1.667$ & $10.667 \pm 1.667$ & $0.000$ & $0.013$ \\
	 & Circle & 3 & No & No & 0 & $7.045 \pm 4.241$ & $8.420 \pm 3.664$ & $14.750 \pm 3.276$ & $12.250 \pm 3.146$ & $24.987$ & $0.029$ \\
	 & Circle & 3 & No & No & 2 & $9.131 \pm 4.288$ & $9.948 \pm 3.936$ & $12.000 \pm 2.191$ & $11.000 \pm 1.817$ & $18.175$ & $0.059$ \\
	 & Circle & 3 & No & Yes & 2 & $5.770 \pm 5.360$ & $5.770 \pm 5.360$ & $12.500 \pm 1.500$ & $12.500 \pm 1.500$ & $0.000$ & $0.018$ \\
	 & Circle & 3 & Yes & No & 0 & $5.563 \pm 3.546$ & $6.643 \pm 3.277$ & $10.200 \pm 3.569$ & $7.600 \pm 3.816$ & $19.995$ & $0.025$ \\
	 & Circle & 3 & Yes & No & 2 & $4.173 \pm 2.654$ & $8.734 \pm 3.423$ & $10.750 \pm 2.496$ & $5.000 \pm 2.121$ & $49.018$ & $0.123$ \\
	 & Circle & 3 & Yes & Yes & 2 & $6.200 \pm 3.143$ & $6.316 \pm 3.087$ & $11.600 \pm 2.804$ & $9.000 \pm 1.924$ & $19.710$ & $0.084$ \\
		\addlinespace
	 & Ellipse & 2 & No & No & 0 & $6.010 \pm 0.958$ & $9.740 \pm 3.056$ & $8.200 \pm 2.653$ & $1.400 \pm 0.400$ & $25.591$ & $0.414$ \\
	 & Ellipse & 2 & No & No & 2 & $5.166 \pm 0.860$ & $7.507 \pm 1.320$ & $12.600 \pm 3.544$ & $1.400 \pm 0.400$ & $29.698$ & $0.562$ \\
	 & Ellipse & 2 & No & Yes & 2 & $5.746 \pm 0.928$ & $6.984 \pm 0.969$ & $11.000 \pm 3.302$ & $1.600 \pm 0.400$ & $17.681$ & $0.391$ \\
	 & Ellipse & 2 & Yes & No & 0 & $4.794 \pm 0.816$ & $6.770 \pm 1.179$ & $8.800 \pm 2.939$ & $1.200 \pm 0.200$ & $24.563$ & $0.283$ \\
	 & Ellipse & 2 & Yes & No & 2 & $5.279 \pm 0.905$ & $8.558 \pm 1.023$ & $10.600 \pm 2.839$ & $1.000 \pm 0.000$ & $36.391$ & $0.420$ \\
	 & Ellipse & 2 & Yes & Yes & 2 & $4.323 \pm 0.885$ & $7.550 \pm 1.321$ & $15.200 \pm 2.922$ & $1.600 \pm 0.600$ & $38.938$ & $0.589$ \\
	 & Ellipse & 3 & No & No & 0 & $6.026 \pm 0.949$ & $6.664 \pm 1.092$ & $11.600 \pm 4.179$ & $1.200 \pm 0.200$ & $8.910$ & $0.363$ \\
	 & Ellipse & 3 & No & No & 2 & $5.958 \pm 1.075$ & $6.648 \pm 1.076$ & $5.800 \pm 2.746$ & $1.400 \pm 0.245$ & $11.515$ & $0.524$ \\
	 & Ellipse & 3 & No & Yes & 2 & $5.869 \pm 0.854$ & $6.886 \pm 0.913$ & $9.800 \pm 3.153$ & $1.600 \pm 0.400$ & $15.468$ & $0.530$ \\
	 & Ellipse & 3 & Yes & No & 0 & $4.752 \pm 0.736$ & $6.780 \pm 1.049$ & $8.800 \pm 2.800$ & $1.000 \pm 0.000$ & $29.121$ & $0.314$ \\
	 & Ellipse & 3 & Yes & No & 2 & $5.350 \pm 1.282$ & $9.062 \pm 2.360$ & $10.800 \pm 4.030$ & $1.000 \pm 0.000$ & $35.413$ & $0.577$ \\
	 & Ellipse & 3 & Yes & Yes & 2 & $4.562 \pm 0.867$ & $7.454 \pm 1.236$ & $14.000 \pm 2.098$ & $1.000 \pm 0.000$ & $35.717$ & $0.554$ \\
\end{tabular}}
\end{table*}

\begin{table*}
\renewcommand\thetable{6}
\centering
\resizebox{\textwidth}{!}{\begin{tabular}{@{}llcccccccccc@{}}
\hline
Model & Shape & \#Ex & Fdbk & SFB & Mem & Best chamf. & Fcham & Steps & Fstep & \% Imp. & \% Semantic \\
\hline
	 \ac{qwen} & Line & 2 & No & No & 0 & $7.592 \pm 1.657$ & $8.548 \pm 2.231$ & $11.000 \pm 2.720$ & $7.400 \pm 3.140$ & $7.364$ & $0.034$ \\
	 & Line & 2 & No & No & 2 & $7.967 \pm 2.032$ & $7.967 \pm 2.032$ & $11.000 \pm 2.915$ & $11.000 \pm 2.915$ & $0.000$ & $0.214$ \\
	 & Line & 2 & No & Yes & 2 & $5.757 \pm 2.652$ & $8.488 \pm 2.728$ & $8.200 \pm 3.338$ & $8.000 \pm 3.162$ & $19.165$ & $0.027$ \\
	 & Line & 2 & Yes & No & 0 & $7.653 \pm 1.741$ & $18.455 \pm 8.155$ & $9.800 \pm 1.772$ & $6.400 \pm 1.208$ & $34.878$ & $0.039$ \\
	 & Line & 2 & Yes & No & 2 & $4.236 \pm 2.421$ & $13.285 \pm 3.206$ & $11.000 \pm 3.209$ & $5.200 \pm 1.356$ & $59.558$ & $0.466$ \\
	 & Line & 2 & Yes & Yes & 2 & $9.153 \pm 7.081$ & $18.403 \pm 8.191$ & $9.800 \pm 2.245$ & $2.800 \pm 0.800$ & $45.798$ & $0.639$ \\
	 & Line & 3 & No & No & 0 & $7.081 \pm 2.225$ & $7.421 \pm 2.433$ & $9.400 \pm 2.581$ & $9.400 \pm 2.581$ & $2.468$ & $0.046$ \\
	 & Line & 3 & No & No & 2 & $5.088 \pm 1.761$ & $11.248 \pm 2.739$ & $11.200 \pm 2.615$ & $4.800 \pm 1.281$ & $37.601$ & $0.590$ \\
	 & Line & 3 & No & Yes & 2 & $15.003 \pm 8.797$ & $16.356 \pm 8.538$ & $12.200 \pm 1.800$ & $11.200 \pm 2.518$ & $10.892$ & $0.137$ \\
	 & Line & 3 & Yes & No & 0 & $6.913 \pm 1.656$ & $15.881 \pm 8.458$ & $11.200 \pm 3.426$ & $3.200 \pm 1.020$ & $35.913$ & $0.058$ \\
	 & Line & 3 & Yes & No & 2 & $8.924 \pm 7.667$ & $14.021 \pm 8.520$ & $14.800 \pm 1.655$ & $3.400 \pm 0.927$ & $41.450$ & $0.551$ \\
	 & Line & 3 & Yes & Yes & 2 & $2.268 \pm 0.919$ & $20.053 \pm 9.575$ & $16.200 \pm 1.685$ & $5.000 \pm 2.121$ & $83.994$ & $0.583$ \\
\addlinespace
	 & Naca & 2 & No & No & 0 & $1.285 \pm 0.487$ & $1.888 \pm 0.356$ & $11.600 \pm 2.421$ & $1.000 \pm 0.000$ & $41.059$ & $0.689$ \\
	 & Naca & 2 & No & No & 2 & $1.297 \pm 0.467$ & $1.953 \pm 0.408$ & $9.200 \pm 2.154$ & $1.000 \pm 0.000$ & $40.370$ & $0.833$ \\
	 & Naca & 2 & No & Yes & 2 & $1.323 \pm 0.508$ & $1.818 \pm 0.472$ & $7.600 \pm 3.187$ & $1.000 \pm 0.000$ & $32.664$ & $0.731$ \\
	 & Naca & 2 & Yes & No & 0 & $1.129 \pm 0.476$ & $2.320 \pm 0.518$ & $11.800 \pm 2.131$ & $1.000 \pm 0.000$ & $57.917$ & $0.477$ \\
	 & Naca & 2 & Yes & No & 2 & $0.887 \pm 0.223$ & $1.953 \pm 0.408$ & $14.800 \pm 2.478$ & $1.000 \pm 0.000$ & $55.968$ & $0.553$ \\
	 & Naca & 2 & Yes & Yes & 2 & $0.889 \pm 0.263$ & $2.677 \pm 0.586$ & $15.200 \pm 2.375$ & $1.000 \pm 0.000$ & $63.523$ & $0.602$ \\
	 & Naca & 3 & No & No & 0 & $1.306 \pm 0.473$ & $1.812 \pm 0.472$ & $9.600 \pm 1.364$ & $1.200 \pm 0.200$ & $32.670$ & $0.651$ \\
	 & Naca & 3 & No & No & 2 & $1.111 \pm 0.283$ & $1.893 \pm 0.356$ & $9.800 \pm 3.216$ & $1.000 \pm 0.000$ & $43.237$ & $0.840$ \\
	 & Naca & 3 & No & Yes & 2 & $1.380 \pm 0.393$ & $1.899 \pm 0.407$ & $4.600 \pm 2.015$ & $1.000 \pm 0.000$ & $31.158$ & $0.802$ \\
	 & Naca & 3 & Yes & No & 0 & $1.007 \pm 0.382$ & $2.130 \pm 0.404$ & $7.400 \pm 1.749$ & $1.000 \pm 0.000$ & $55.782$ & $0.530$ \\
	 & Naca & 3 & Yes & No & 2 & $0.876 \pm 0.195$ & $1.985 \pm 0.436$ & $13.800 \pm 2.800$ & $1.000 \pm 0.000$ & $56.462$ & $0.742$ \\
	 & Naca & 3 & Yes & Yes & 2 & $1.080 \pm 0.401$ & $1.980 \pm 0.437$ & $13.000 \pm 2.214$ & $1.000 \pm 0.000$ & $52.116$ & $0.812$ \\
		\addlinespace
	 & Parabola & 2 & No & No & 0 & $890.798 \pm 24.891$ & $896.765 \pm 23.919$ & $7.200 \pm 2.746$ & $1.200 \pm 0.200$ & $0.672$ & $0.594$ \\
	 & Parabola & 2 & No & No & 2 & $807.431 \pm 18.711$ & $897.205 \pm 24.825$ & $13.600 \pm 2.135$ & $1.000 \pm 0.000$ & $9.804$ & $0.810$ \\
	 & Parabola & 2 & No & Yes & 2 & $837.704 \pm 21.792$ & $897.545 \pm 24.517$ & $10.600 \pm 3.655$ & $1.200 \pm 0.200$ & $6.526$ & $0.723$ \\
	 & Parabola & 2 & Yes & No & 0 & $847.932 \pm 38.709$ & $898.009 \pm 24.439$ & $10.000 \pm 2.702$ & $1.800 \pm 0.583$ & $5.732$ & $0.342$ \\
	 & Parabola & 2 & Yes & No & 2 & $814.122 \pm 39.460$ & $897.491 \pm 24.524$ & $9.200 \pm 2.922$ & $1.000 \pm 0.000$ & $8.900$ & $0.557$ \\
	 & Parabola & 2 & Yes & Yes & 2 & $754.452 \pm 54.486$ & $899.164 \pm 26.358$ & $18.200 \pm 0.860$ & $1.200 \pm 0.200$ & $16.416$ & $0.819$ \\
	 & Parabola & 3 & No & No & 0 & $888.611 \pm 24.141$ & $896.433 \pm 23.917$ & $10.800 \pm 2.764$ & $1.000 \pm 0.000$ & $0.874$ & $0.600$ \\
	 & Parabola & 3 & No & No & 2 & $848.301 \pm 41.711$ & $897.992 \pm 24.505$ & $13.400 \pm 3.219$ & $1.200 \pm 0.200$ & $5.692$ & $0.613$ \\
	 & Parabola & 3 & No & Yes & 2 & $895.240 \pm 25.034$ & $897.025 \pm 24.547$ & $4.200 \pm 3.200$ & $1.000 \pm 0.000$ & $0.205$ & $0.598$ \\
	 & Parabola & 3 & Yes & No & 0 & $884.636 \pm 25.279$ & $897.639 \pm 25.247$ & $11.200 \pm 1.497$ & $1.200 \pm 0.200$ & $1.452$ & $0.430$ \\
	 & Parabola & 3 & Yes & No & 2 & $818.250 \pm 35.861$ & $897.503 \pm 24.680$ & $12.600 \pm 3.444$ & $1.200 \pm 0.200$ & $8.706$ & $0.569$ \\
	 & Parabola & 3 & Yes & Yes & 2 & $709.656 \pm 67.207$ & $897.195 \pm 23.691$ & $15.000 \pm 2.470$ & $1.000 \pm 0.000$ & $21.299$ & $0.840$ \\
\end{tabular}}
\caption{Performance results across shapes and settings, where $\pm$ denotes the standard error.}
\end{table*}
}

\section{Appendix: Memory dynamics}
\label{app:memory_dynamics}

The impact of memory and the number of retrieved design examples depends on both architecture and scale. \ac{qwen}, a compact model with step-by-step reasoning training, benefits consistently from increasing memory (Mem $0\!\to\!2$), with aggregate means falling by $\approx 11.6$ at 2 examples ($150.498\!\to\!138.882$) and by $\approx 11.4$ at 3 examples ($152.391\!\to\!141.034$), with the largest impact on the most challenging shape (Parabola). By contrast, \ac{llama}, a large dense model, shows a memory--examples interaction: memory helps at 2 examples ($147.819\!\to\!143.836$, $\Delta\!\approx\!-3.98$) but can harm at 3 examples ($142.515\!\to\!144.959$, $\Delta\!\approx\!+2.44$), suggesting that very large models are more sensitive to the volume of contextual examples (additional examples can induce interference). \ac{gemma} displays only modest, variable responses to memory and example count (e.g., 2 examples, Mem $0\!\to\!2$: $153.267\!\to\!153.038$, $\Delta\!\approx\!-0.23$).

\paragraph{Architecture and scale.} Across the evaluated shapes, model family and training regime were more predictive of performance than raw parameter count. The \ac{lrm}-style model \ac{qwen} achieved the best aggregate minimisation, largely driven by its superior performance on the Parabola task, whereas the dense \ac{llama} excelled on the other shapes (Circle, Ellipse, Line). Averaged over conditions, Parabola means were \ac{qwen} $=833.09$, \ac{llama} $=845.99$, \ac{gemma} $=883.79$; for the other shapes \ac{llama} attained the lowest means (Circle $=5.00$, Ellipse $=4.06$, Line $=3.54$). On NACA and \ac{lb}, the models were close ($\approx 1.13$ and $\approx 8.1$--$8.2$ for \ac{qwen} and \ac{llama}), with \ac{gemma} uniformly higher. All models benefit from the combination of SFB and external feedback, but the magnitude varies: \ac{qwen} exhibits the largest absolute improvements under SFB+feedback, \ac{llama} moderate gains, and \ac{gemma} the smallest --- consistent with the view that step-by-step reasoning training enables better exploitation of structured symbolic feedback.

\section{Appendix: Memory --- formal description and learning dynamics}
\label{app:memory_dynamics_method}

This appendix expands the memory component summarised in Section~\ref{sec:method}.

\paragraph{Formal description.} Let $M_t = \{m_1, m_2, \dots, m_n\}$ denote the set of stored mechanisms at iteration $t$. A mechanism $m_i$ is stored only if it satisfies the execution criterion $\mathcal{V}(m_i) = 1$. Retrieval follows
$m_{\text{retrieved}} = \arg\min_{m_i \in M_t} d(m_i, m_{\text{target}})$,
where $d(\cdot,\cdot)$ is a predefined distance metric. Memory consists exclusively of mechanisms that executed successfully, which preserves validity across the repository, eliminates invalid designs from future context, and favours convergence by retaining proven solutions.

\paragraph{Retrieval methodology.} Memory retrieval implements a proximity-based selection strategy: (1) mechanisms are prioritised by closeness to the current design objectives; (2) when multiple examples are required, the top-$k$ closest mechanisms are returned.

\paragraph{Progressive specialisation.} As iterations proceed, the memory accumulates designs that successfully navigate the specific constraints of the current problem. Early iterations contribute generic mechanisms satisfying basic kinematic requirements; later iterations add designs that handle edge cases, optimise for specific trajectory features, or employ domain-specific tricks discovered through the critique feedback loop.

\paragraph{Pattern transfer.} The retrieval mechanism ($\arg\min_{m_i \in M_t} d(m_i, m_{\text{target}})$) ensures that the \ac{llm} encounters designs geometrically similar to current objectives --- e.g.\ for a quarter-circle trajectory, it retrieves past designs for circular or arc-shaped paths --- enabling transfer of geometric patterns (e.g.\ link-length ratios for a given curvature) without explicit feature engineering.

\paragraph{Failure avoidance.} The selective storage criterion ($\mathcal{V}(m_i) = 1$) prevents the \ac{llm} from re-exploring regions that violate physical constraints (disconnected link chains, impossible joint configurations), implicitly learning the valid subspace of the design space.

\paragraph{Formal learning objective.} Memory-augmented learning can be cast as minimising expected trajectory error over the task distribution:
$\mathbb{E}_{\mathcal{T} \sim p(\text{tasks})}\!\left[ \min_{t \leq T} d(\mathcal{M}_t, \mathcal{T}) \mid M_t \right]$,
where $\mathcal{M}_t$ is the best mechanism at iteration $t$ and $M_t$ the memory state. Memory acts as a sufficient statistic for task-relevant knowledge, yielding lower expected error with fewer iterations as $|M_t|$ grows.

The paradigm differs from supervised learning (no ground-truth labels), reinforcement learning (no learnable policy parameters), and evolutionary algorithms (no population-based crossover): it is a form of \emph{linguistic program synthesis with geometric grounding} in which memory retrieval and symbolic feedback guide systematic exploration without parameter updates.

\section{Appendix: Generalisability beyond mechanism synthesis}
\label{app:generalisability}

Although we demonstrate the linguistic--numerical interface on planar mechanism synthesis, the underlying principle---providing a structured symbolic translation layer between any simulator's numerical outputs and an \ac{llm}'s linguistic reasoning---is domain-general. Several characteristics make a domain amenable to this interface:

\begin{enumerate}
    \item \textbf{Simulability}: A physics-based or mathematical simulator must exist that can produce numerical trajectories or fields from design parameters. This is satisfied across computational mechanics, molecular dynamics, fluid simulation, and robotic motion planning.
    \item \textbf{Symbolic approximability}: The continuous behaviour must be well-approximated by compact analytical expressions discoverable via symbolic regression. This holds for periodic motions (robot gaits, oscillatory circuits), algebraic curves (molecular bond geometries, structural deformations), and piecewise-smooth trajectories (tool paths, vehicle routes).
    \item \textbf{Iterative refinability}: The design problem must admit progressive refinement, where feedback on one iteration can inform the next. This is broadly true of optimisation problems in engineering and scientific domains.
\end{enumerate}

Concrete domains where the interface could be instantiated include: \emph{molecular geometry optimisation}, where bond lengths, angles, and dihedral angles define continuous spatial configurations that a simulator produces numerically and an \ac{llm} could critique symbolically; \emph{robotic trajectory planning}, where end-effector paths through 3D space bear direct structural similarity to the 2D mechanism trajectories studied here; \emph{structural topology optimisation}, where stress distributions and deformation fields from FEM simulators can be approximated by symbolic surrogates; and \emph{metamaterial design}, where unit-cell geometries determine continuous wave-propagation properties. In each case, the same interface cycle---linguistic specification $\to$ parametric design $\to$ simulation $\to$ \textbf{symbolic interface} $\to$ linguistic critique---could be instantiated with domain-appropriate simulators and symbolic regression targets.

\section{Appendix: Broader connections}
\label{app:broader_connections}

The interface paradigm connects to several active research threads. First, it extends the \emph{tool-augmented language model} paradigm by showing that tools (here, simulators and symbolic regression) can serve not just as information sources but as \emph{interface components} that translate between representational domains, expanding the class of problems amenable to linguistic reasoning. Second, the method instantiates a form of \emph{neuro-symbolic integration} where the symbolic component (SR expressions) is not hard-coded but discovered from data, creating a dynamic interface layer that adapts to each design problem. Third, the memory-based learning component relates to \emph{retrieval-augmented generation}, but with a crucial difference: the retrieved content is not static knowledge but dynamically accumulated problem-solving experience generated through the interface, making the method a form of experience-augmented reasoning.

\end{document}